\documentclass[lettersize,journal]{IEEEtran}
\usepackage{amsmath,amsfonts}
\usepackage{algorithm}
\usepackage{array}
\usepackage[caption=false,font=footnotesize,labelfont=sf,textfont=sf]{subfig}
\usepackage{textcomp}
\usepackage{stfloats}
\usepackage{url}
\usepackage{verbatim}
\usepackage{graphicx}
\usepackage{cite}
\def\Vbar{\perp\!\!\!\perp}
\usepackage{hyperref}
\usepackage{xcolor}

\usepackage{amsthm}
\usepackage{fdsymbol}
\usepackage{float}
\usepackage{tabularx}
\usepackage{multirow}
\usepackage{booktabs}

\theoremstyle{plain}
\usepackage{newfloat}
\usepackage{listings}
\usepackage{algpseudocode}

\usepackage{microtype}      

\usepackage[normalem]{ulem}

\newtheorem{definition}{Definition}
\newtheorem{lemma}{Lemma}
\newtheorem{theorem}{Theorem}
\newtheorem{corollary}{Corollary}

\newtheorem{assumption}{Assumption}

\def\bR{{\mathbb R}}
\def\bE{{\mathbb E}}

\begin{document}

\title{Nonparametric Heterogeneous Long-term Causal Effect Estimation via Data Combination}


\author{Weilin Chen, 
        Ruichu~Cai\textsuperscript{*},~\IEEEmembership{Senoir Member,~IEEE,}
        Junjie Wan,
        Zeqin Yang,
        José Miguel Hernández-Lobato
\thanks{Weilin Chen is with the School of Computer Science, Guangdong University of Technology, Guangzhou, 510006, China (e-mail: chenweilin.chn@gmail.com).}
\thanks{Ruichu Cai is with the School of Computer Science, Guangdong University of Technology, Guangzhou, 510006, China and Peng Cheng Laboratory, Shenzhen, China (e-mail: cairuichu@gmail.com).}
\thanks{Junjie Wan and Zeqin Yang are with the School of Computer Science, Guangdong University of Technology, Guangzhou, 510006, China (e-mail: wj1205131700@gmail.com and youngzeqin@gmail.com).}
\thanks{José Miguel Hernández-Lobato is with the Department of Engineering,
University of Cambridge, Cambridge CB2 1PZ, United Kingdom (e-mail: jmh233@cam.ac.uk).}
\thanks{Corresponding author: Ruichu Cai.} 
        
}

\markboth{Journal of \LaTeX\ Class Files,~Vol.~14, No.~8, August~2021}%
{Shell \MakeLowercase{\textit{et al.}}: A Sample Article Using IEEEtran.cls for IEEE Journals}


\maketitle

\begin{abstract}
Long-term causal inference has drawn increasing attention in many scientific domains.
Existing methods mainly focus on estimating average long-term causal effects by combining long-term observational data and short-term experimental data.
However, it is still understudied how to robustly and effectively estimate heterogeneous long-term causal effects, significantly limiting practical applications.
In this paper, we propose several two-stage style nonparametric estimators for heterogeneous long-term causal effect estimation, including propensity-based, regression-based, and multiple robust estimators.
We conduct a comprehensive theoretical analysis of their asymptotic properties under mild assumptions, with the ultimate goal of building a better understanding of the conditions under which some estimators can be expected to perform better.
Extensive experiments across several semi-synthetic and real-world datasets validate the theoretical results and demonstrate the effectiveness of the proposed estimators.
\end{abstract}


\section{Introduction}

\begin{figure}[t] 
	\centering
	\subfloat[Experimental data ($G=E$)]{
		\includegraphics[width=0.24\textwidth]{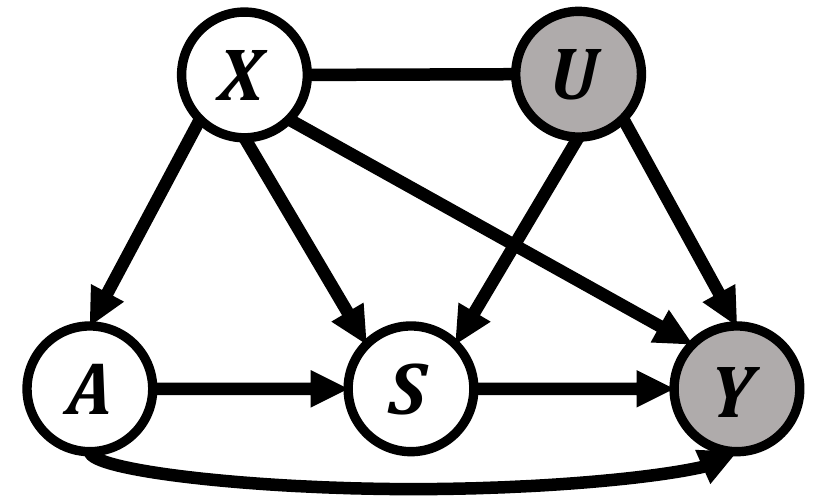} 
		\label{figure exp}
	}
	\subfloat[Observational data ($G=O$)]{
		\includegraphics[width=0.24\textwidth]{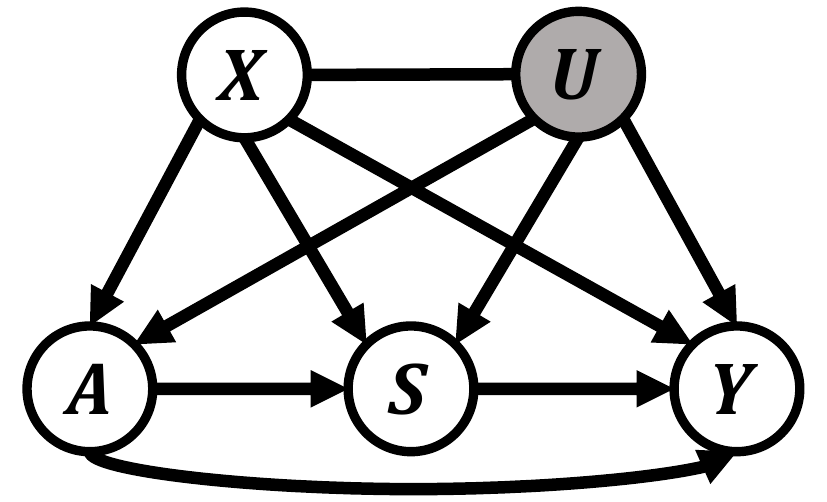}
		\label{figure obs}
	}
	\caption{Causal graphs of experimental data and observational data with $X$ being covariates, $U$ being unobserved confounders, $A$ being treatment, $S$ being short-term outcome, and $Y$ being long-term outcome. Gray nodes denote unobserved variables and white nodes denote observed variables. Arrows denote causal relationships. 
    Fig. \ref{figure exp} represents the causal graph of the short-term experimental data,
    where treatment $A$ is not affected by unobserved confounders $U$ and the long-term outcome $Y$ is unobserved.
    Fig. \ref{figure obs} represents the causal graph of the long-term observational data, 
    where the unobserved confounders $U$ affect treatments $A$ and outcome $S,Y$ and the long-term outcome $Y$ can be observed..}
	 \label{fig: causal graph} 
\end{figure}

\IEEEPARstart{L}{ong}-term causal effect estimation has drawn increasing attention in many scientific areas, such as medicine \cite{fleming1994surrogate} and advertising \cite{hohnhold2015focusing}. 
However, since conducting long-term experiments is not feasible due to the high cost, many studies seek to combine \textbf{short-term experiential data} and \textbf{long-term observational data} to estimate long-term effects \cite{athey2019surrogate,athey2020combining,ghassami2022combining,cai2024long,yang2024estimating}.
The typical causal graphs are shown in Fig. \ref{fig: causal graph}, where the long-term causal effects are not identifiable only using the experimental data due to the missingness of long-term outcome $Y$, as well as the observational data due to the unobserved confounders $U$. 
Therefore, a natural question is how to combine two different types of data for long-term causal inference.

Existing methods explore various assumptions to fuse experimental data and observational data to estimate long-term causal effects.
A widely used assumption is the Latent Unconfoundedness (LU) \cite{athey2020combining, chen2023semiparametric}.
LU assumes that in the observational data, the short-term outcomes $S$ can totally mediate the causal path from treatment $A$ to long-term potential outcome $Y(a)$, which graphically rules out the causal edge from $U$ to $Y$, indicating the unobserved confounder can only affect treatment $A$ and short-term outcomes $S$.
To allow the existence of causal edge $U \rightarrow Y$, AmirEmad et~al. \cite{ghassami2022combining} propose the Conditional Additive Equi-Confounding Bias (CAECB) assumption, which requires the short-term confounding bias to be equal to the long-term one.
Under CAECB assumption, AmirEmad et~al. \cite{ghassami2022combining} further propose an influence function-based estimator for long-term average causal effects.

Existing methods \cite{athey2020combining, ghassami2022combining, chen2023semiparametric}, however, mainly focus on identifying and estimating average long-term effects, 
which can not be directly extended to Heterogeneous Long-term Causal Effects (HLCE), significantly limiting their practical utility and broader applicability.
In many real-world applications, understanding HLCE is essential for designing personalized strategies tailored to individual needs, rather than relying on average effects that may not account for the diverse heterogeneity across different individuals.
For example, in medical treatments, patients often present with varying conditions and responses (heterogeneity), necessitating personalized treatment plans to effectively improve their long-term recovery outcomes.
Consequently, the lack of heterogeneity consideration in existing methods restricts the potential for delivering interventions specifically designed for individuals.

In this paper, to fill such a research gap, we focus on designing the HLCE estimators and providing an extensive theoretical analysis of the asymptotical behaviors of our proposed estimators.
Specifically, we propose several HLCE estimators under the CAECB assumption within a two-stage regression framework, which are model-agnostic algorithms that decompose the task of estimating HLCE into multiple sub-problems, each solvable using any supervised learning/regression methods.
The most important baseline method that we propose is a Multiple Robust (MR) estimator, which is shown to be consistent in the union of four different model specifications. 
This is different from existing double/multiple robust methods for long-term inference \cite{chen2023semiparametric,ghassami2022combining}, which only show the robust property in terms of average effects.
In the theoretical part of this paper, we analyze the convergence rates of the proposed methods within a generic nonparametric regression framework, showing why a baseline estimator may outperform others and how the MR property is achieved.
In our practical part, we leverage our theoretical results and, on top of it, build neural network-based HLCE estimators, utilizing the shared representation technique proposed by Johansson et~al. \cite{johansson2022generalization}.
Overall, our contribution can be summarized as follows:
\begin{itemize}
    \item We study the problem of heterogeneous long-term effect estimation under the Conditional Additive Equi-Confounding Bias assumption and design several two-stage baseline methods, which use unbiased pseudo outcome regression based on outcome regression and inverse propensity weighting.
    \item We further propose a multiple robust estimator of heterogeneous long-term effects, 
    which shows attractive properties in terms of model misspecification and convergence rates.
    \item We provide an extensive theoretical analysis of the convergence rates of the proposed baseline estimators and the MR estimator. 
    Extensive experimental studies, conducted on multiple synthetic and semi-synthetic datasets, demonstrate the correctness and effectiveness of our proposed method.
\end{itemize}

\section{Related Work}  

\textbf{Long-term Causal Inference}
For decades, many studies have explored the validity of a surrogate, i.e., what kind of short-term outcomes can reliably predict long-term causal effects. 
Various criteria are proposed for a valid surrogate, e.g., prentice criteria \cite{prentice1989surrogate}, principal criteria \cite{frangakis2002principal}, strong surrogate criteria \cite{lauritzen2004discussion}, causal effect predictiveness \cite{gilbert2008evaluating}, and consistent surrogate and its variants \cite{chen2007criteria, ju2010criteria, yin2020novel}.
Recently, many works have studied estimating long-term causal effects based on surrogates. 
One prominent line of research assumes the unconfoundedness assumption.
Under the unconfoundedness assumption, LTEE \cite{cheng2021long} and Laser \cite{cai2024long} are based on different designed neural networks for long-term causal inference.
EETE \cite{kallus2020role} studies the data efficiency from the surrogate and proposes efficient estimation for treatment effect.
ORL \cite{tran2023inferring} proposes a doubly robust estimator for average treatment effects with only short-term experiments, additionally assuming stationarity conditions between short and long-term outcomes.
\cite{wu2024policy} proposes a policy learning method for balancing short-term and long-term rewards.
\underline{Different} from these works, we do not assume the unconfoundedness assumption, and we use the data combination technique to solve the problem of unobserved confounders.
Another line of research, which also avoids the unconfoundedness assumption, tackles the issue by combining experimental and observational data — a setting known as data combination.
This setting is initialized by the method proposed by Athey et~al. \cite{athey2019surrogate}, 
which, under surrogacy assumption, constructs the so-called Surrogate Index as the substitutions for long-term outcomes in the experimental data to achieve effect identification.
As follow-up work, \cite{athey2020combining} assumes latent unconfoundedness assumption, i.e., short-term potential outcomes can mediate the long-term potential outcomes, to identify long-term causal effects.
Other feasible assumptions \cite{ghassami2022combining}  are proposed to replace the latent unconfoundedness assumption, e.g., the additive equi-confounding bias assumption.
Based on proximal methods, the sequential structure surrogates are studied \cite{imbens2022long}.
Learn \cite{yang2024estimating} proposes a reweighting schema to align observational data and experimental data, enabling effect identification.
\underline{However}, these works mostly focus on the average treatment effects or do not consider double/multiple robust estimators for heterogeneous causal effects.
Different from these works above, we address the overlooked problem by providing several heterogeneous long-term causal effect estimators, including regression-based, propensity score-based, and multiple robust estimators, and provide a comprehensive theoretical analysis of their properties.

\textbf{Double/Multiple Robustness}
A double/multiple Robust estimator is an estimator that remains consistent when part of nuisance functions are inconsistent.
Regarding average treatment effect estimation, the most well-known estimator is the augmented inverse propensity weighted (AIPW) estimator \cite{robins1994estimation} in the traditional scenario. 
AIPW consists of a regression model and a propensity model \cite{rosenbaum1983central}, and it is consistent as long as one of the models is consistent.
Similarly, doubly robust estimators for average causal effects are proposed in various scenarios.
\cite{singh2024double} and \cite{wang2018bounded} propose a doubly and multiple robust estimator, respectively, for average causal effect in the instrumental variable (IV) setting.
\cite{tchetgen2012semiparametric} proposes a multiple robust estimator for mediation analysis.
For continuous average effect estimation, \cite{kennedy2017non} proposes a nonparametric estimator leveraging kernel methods.
More related to our work, \cite{ghassami2022combining} proposes a multiple robust estimator for long-term average effects in the same setting as ours. 
\underline{However}, these works above are not applicable to estimate the heterogeneous effects.
Different from them, our work focuses on designing multiple robust heterogeneous effect estimator instead of average effect estimators.
Additionally, many works also study the double/multiple robust estimator for heterogeneous effects.
\cite{kennedy2023towards} analyzes the doubly robust estimator in the standard setting and derives doubly robust convergence rates.
\cite{frauenestimating} extends to the IV setting and proposes a corresponding multiple robust estimator.
\underline{However}, the multiple robust estimation for long-term heterogeneous effects is still an understudied problem. 
In this paper, we propose a multiple robust heterogeneous effect estimator based on neural networks and also provide a detailed theoretical analysis.

\section{Problem Definition, Assumptions}

Let 
$A\in \{0,1\}$ be the treatment variable, 
$X\in \bR^{d}$ be the observed covariates where $d$ is the dimension of $X$,
$U\in \bR^{d_U}$ be the unobserved covariates,
$S$ be the short-term outcome variable,
and $Y$ be the long-term outcome variable.
Further, we denote
$S(a)\in \bR$ as the potential short-term outcome variable 
and $Y(a)\in \bR$ as the potential long-term outcome variable.
Following \cite{athey2020combining,  chen2023semiparametric, ghassami2022combining, imbens2022long}, 
we denote $G \in \{E,O\}$ as the indicator of data source, where $G=E$ indicates the experimental data and $G=O$ indicates the observational data. 
Let lowercase letters (e.g., $a, x, u, s,y, s(a),y(a)$) denote the value of the above random variables.
Let the index $i$ denote a specific unit, e.g., $x_i$ is the covariate value of unit $i$. 
Then, the experimental data and the observational data are denoted as $\mathbb D_e =\{a_i, x_i, s_i, G_i=E\}_{i=1}^{n_e}$ and $\mathbb D_o = \{a_i, x_i, s_i, y_i, G_i=O\}_{i=n_e+1}^{n_e+n_o}$, where $n_e$ and $n_o$ are the size of experimental data and the observational data respectively. 

\textbf{Task}: Given a short-term experimental dataset $\{a_i, x_i, s_i, G_i=E\}_{i=1}^{n_e}$ and a long-term observational dataset $\{a_i, x_i, s_i, y_i, G_i=O\}_{i=n_e+1}^{n_e+n_o}$, the estimand in this paper is the Heterogeneous Long-term Causal Effects (HLCE):
\begin{equation}
    \tau (x) = \bE [Y(1)-Y(0)|X=x].
\end{equation}
Here, $\tau(x)$ represents the difference between the long-term outcome $Y$ of the specific unit $x$ when treated and the same unit when not treated (control). 
$\tau(x)$ provides valuable insights into how the treatment impacts long-term outcomes, facilitating the design of personalized strategies in various applications.
However, $\tau (x)$ can not be identified without further assumptions, 
since the experimental data lacks the long-term outcome $Y$ and the observational data suffers from the latent confounding problem.
To ensure the identification of long-term effects, we make the following assumptions throughout this paper:

\begin{assumption}[Consistency] \label{assum: consist}
If a unit is assigned treatment, we observe its associated potential outcome. 
Formally, if $A=a$, then $Y=Y(a), S=S(a)$.
\end{assumption}

\begin{assumption}[Positivity] \label{assum: positi}
The treatment assignment is non-deterministic. 
Formally, $\forall a,x$, we have $0<P(A=a|X=x)<1,~ 0<P(G=O|A=a, X=x)<1$.
\end{assumption}

\begin{assumption} [Weak internal validity of observational data] \label{assum: internal validity of obs}
Unobserved confounders exist in Observational data. 
Formally, $\forall a \in \{0,1\}$, $A\Vbar \{Y(a),S(a)\}|X, U, G=O$ and $A \not{\Vbar} \{Y(a),S(a)\}|X, G=O$.
\end{assumption}

\begin{assumption} [Internal validity of experimental data] \label{assum: internal validity of exp}
There are no unobserved confounders in experimental data.
Formally, $\forall a \in \{0,1\}$, $A\Vbar \{Y(a),S(a)\}|X, G=E$.
\end{assumption}

\begin{assumption} [External validity of experimental data] \label{assum: external validity of exp}
The distribution of the potential outcomes is invariant to whether the data belongs to the experimental or observational data. Formally,
    $\forall a \in \{0,1\}$, $G\Vbar \{Y(a),S(a)\}|X$.
\end{assumption}

\begin{assumption} [Conditional Additive Equi-Confounding Bias, CAECB] \label{assum: equ bias}
The difference of conditional expected value of short-term potential outcomes across treated and control groups is the same as that of the long-term potential outcome variable.
Formally, $\forall a$, we have
\begin{equation}
\begin{aligned}
    & \bE [S(a)|X,A=0,G=O]-\bE[S(a)|X,A=1,G=O]
    \\ = & \bE [Y(a)|X,A=0,G=O]-\bE[Y(a)|X,A=1,G=O].
\end{aligned}
\end{equation}
\end{assumption}

The causal graphs of observational data and experimental data satisfying the above assumptions are shown in Fig. \ref{fig: causal graph}.
Assumptions \ref{assum: consist} and \ref{assum: positi} are standard assumptions in causal inference \cite{rubin1974estimating, imbens2000role}. 
Assumptions \ref{assum: internal validity of obs}, \ref{assum: internal validity of exp} and \ref{assum: external validity of exp} are mild and widely assumed in data combination settings \cite{shi2023data, imbens2022long, athey2019surrogate, athey2020combining, hu2023identification}. 
Specifically, Assumption \ref{assum: internal validity of obs} allows the existence of latent confounders in observational data, thus it is much weaker than the traditional unconfoundedness assumption.
Assumption \ref{assum: internal validity of exp} is reasonable and can be achieved since the treatment assignment mechanism is under control in the experiments.
Assumption \ref{assum: external validity of exp} connects the potential outcome distributions between observational and experimental data.
Most importantly, Assumption \ref{assum: equ bias}, proposed by \cite{ghassami2022combining}, ensures confounding biases conditional on covariates $X$ are equal between short-term and long-term outcomes. This assumption offers a route to identify long-term unobserved confounding and further identify long-term effects.

Under the assumptions above, the heterogeneous causal effects can be identified, as stated in the following theorem.

\begin{theorem} \label{theo: identifi} Suppose Assumptions \ref{assum: consist}, \ref{assum: positi}, \ref{assum: internal validity of obs}, \ref{assum: internal validity of exp}, \ref{assum: external validity of exp} and \ref{assum: equ bias} hold, then $\tau(x)$ can be identified as follows:
    \begin{equation} \label{eq: identi} \small
      \begin{aligned}
       & \tau(x) \\
            = & \mathbb E[Y(1)|X=x,G=O] - \mathbb E[Y(0)|X=x,G=O] \\
            = & \mathbb E[Y|X=x,G=O,A=1] - \mathbb E[Y|X=x,G=O,A=0]  \\
            & +  \mathbb E[S|X=x,G=E,A=1] - \mathbb E[S|X=x,G=E,A=0] \\
            &  + \mathbb E[S|X=x,G=O,A=0] - \mathbb E[S|X=x,G=O,A=1].
      \end{aligned}
    \end{equation}
\end{theorem}

Proof can be found in Appendix \ref{app: proof of iden}. A similar identification result in terms of average causal effects $\bE[Y(1)-Y(0)]$ has been shown in \cite{ghassami2022combining}. Different from them, we establish the identification result in terms of heterogeneous causal effects. More importantly, we focus on the estimation of $\tau(x)$ in this paper and propose several baseline estimators and a multiple robust estimator as shown in the following sections.

\section{Heterogenoeous Long-term Effect Estimators}
\label{sec: estimator}

In this section, we focus on the estimation of HLCE $\tau(x)$.
To begin with, motivated by the identification results in Theorem \ref{theo: identifi}, we design regression-based and propensity-based estimators of HLCE $\tau(x)$, which is shown to be consistent with correctly specified nuisance functions.
Further, we design a multiple robust estimator of $\tau(x)$ by combining regression-based and propensity-based estimators. This estimator shows a more appealing property, which achieves consistency as long as only one of four sets of nuisance functions is correctly specified.

To be precise and convenient, we denote several nuisance functions that will be used in this paper as follows:
\begin{equation} \label{defined nuisance}
    \begin{aligned}
      &  \mu_S^E(A,X) = \bE [S|A,X,G=E], \\
      &  \mu_S^O(A,X) = \bE [S|A,X,G=O], \\
      &  \mu_Y^O(A,X) = \bE [Y|A,X,G=O], \\
      &  \pi^E(X) = \bE [A=1|X,G=E], \\
      &  \pi^O(X) = \bE [A=1|X,G=O], \\
      &  \pi^G(X) = \bE [G=E|X] .
    \end{aligned}
\end{equation}

\subsection{Baselines: Two-stage Regression and Propensity Estimator}

Directly following Eq. \eqref{eq: identi}, we can design an one-stage regression-based naive estimator $\hat \tau_{naive}(x)$ as $\hat \tau_{naive}(x)=\hat \mu_Y^O(1,x) - \hat \mu_Y^O(0,x) + \hat \mu_S^E(1,x) - \hat \mu_S^E(0,x) + \hat \mu_S^O(0,x) - \hat \mu_S^O(1,x)$, which is consistent with correctly specified nuisance functions $\mu_S^E(a,x),\mu_S^O(a,x),$ and $\mu_Y^O(a,x)$. However, the propensity score-based estimator can not be directly applied to estimate heterogeneous causal effects, since it is designed to estimate average causal effects.
To extend the propensity score-based estimator to estimate HLCE, we propose a two-stage propensity-based estimator, denoted as $\hat \tau_{pro}(x)$.
To be consistent, we also consider a similar kind of two-stage regression for the regression-based estimator, denoted as $\hat \tau_{reg} (x)$ and we also provide the similar properties between one-stage and two-stage regression-based estimators in Section \ref{sec: converg rate}.

The two-stage estimators follow a two-step process: (1) fitting the nuisance functions, and (2) regressing a pseudo outcome $\hat Y$ (constructed using the nuisance functions) on $X$ to obtain $ \hat \tau(x)$. 
For the second-stage $\hat \tau(x)$ to be unbiased, the designed pseudo outcomes $\hat Y$ should satisfy $\bE [\hat Y|X=x]=\tau(x)$.
Therefore, motivated by the outcome regression model and the inverse propensity weighting model, we design two different pseudo outcomes $\hat Y_{reg}$ and $\hat Y_{pro}$, resulting in two unbiased estimators $\hat \tau_{reg} (x)$ and $\hat \tau_{pro} (x)$ for HLCE, respectively.

Specifically, the regression-based estimator $\hat \tau_{reg} (x)$ is constructed by:
\begin{itemize}
    \item [S1.] Fitting nuisance functions $\hat \mu_S^E(a,x)$, $\hat \mu_S^O(a,x)$, and $\hat \mu_Y^O(a,x)$;
    \item [S2.] Regressing the pseudo outcome $\hat Y_{reg}$ on covariates $X$ to obtain $\hat \tau_{reg} (x)$, i.e., $\hat \tau_{reg} (x)=\hat \bE [\hat Y_{reg}|X=x]$, where the pseudo outcome $\hat Y_{reg}$ follows
    \begin{equation} 
    \begin{aligned}
    & \hat Y_{reg} \\
    = & \mathbb I (G=O) \left[ (-1)^{1-A} \left( Y - \hat \mu_Y^O(1-A,X)   \right. \right. \\
    & \left. \left. - S + \hat \mu_S^O(1-A,X) \right)  +\hat \mu_S^E(1,X) - \hat \mu_S^E(0,X) \right] \\
     & + \mathbb I (G=E) \left[ (-1)^{1-A} \left(S - \hat \mu_S^E(1-A,X) \right) \right.  \\
     &  \left. +\hat  \mu_Y^O(1,X) - \hat \mu_Y^O(0,X) + \hat \mu_S^O(0,X)
      - \hat \mu_S^O(1,X)) \right] .
    \end{aligned}
    \end{equation}
\end{itemize}
Similarly, the propensity-based estimator $\hat \tau_{pro}(x)$ is constructed by:
\begin{itemize}
    \item [S1.] Fitting nuisance functions $\hat \pi^E(a,x)$, $\hat \pi^O(a,x)$, and $\hat \pi^G(a,x)$;
    \item [S2.] Regressing the pseudo outcome $\hat Y_{pro}$ on covariates $X$ to obtain $\hat \tau_{pro} (x)$, i.e., $\hat \tau_{pro} (x)=\hat \bE [\hat Y_{pro}|X=x]$, where the pseudo outcome $\hat Y_{pro}$ follows
    \begin{equation} 
    \begin{aligned}        
    & \hat Y_{pro} \\
    = & \frac{(-1)^{1-A}}{1-A+(-1)^{1-A} \hat \pi ^E(X)} \frac{\mathbb I(G=E)}{p(G=O)}
     (\frac{1}{\hat \pi ^G(X)}-1)S \\
    & + \frac{\mathbb I(G=O)}{p(G=O)}\frac{(-1)^{1-A}}{1-A+(-1)^{1-A} \hat \pi ^O(X)} (Y-S)   .
    \end{aligned}
    \end{equation}
\end{itemize}

Such two-stage estimators can be implemented by any off-the-shelf machine learning methods, e.g., kernel regressions and neural networks. We provide their consistency results in the following lemma.

\begin{lemma} [Baselines Consistency] \label{lem: baselines consis}
    Suppose Assumptions \ref{assum: consist}, \ref{assum: positi}, \ref{assum: internal validity of obs}, \ref{assum: internal validity of exp}, \ref{assum: external validity of exp} and \ref{assum: equ bias} hold, then $\hat \tau_{reg}(x)$ is consistent if the nuisance functions $\hat \mu_S^E(a,x)$, $\hat \mu_S^O(a,x)$, and $\hat \mu_Y^O(a,x)$ are consistent, and similarly  $\hat \tau_{pro}(x)$ is consistent if the nuisance functions $\hat \pi^E(x)$, $\hat \pi^O(x)$, and $\hat \pi^G(x)$ are consistent.
\end{lemma}

Proof can be found in Appendix \ref{app: proof of consi}.
Lemma \ref{lem: baselines consis} promises the correctness of our two baseline estimators $\hat \tau_{reg}(x)$ and $\hat \tau_{pro}(x)$. 
The consistency of $\hat \tau_{reg}(x)$ and $\hat \tau_{pro}(x)$ requires their used nuisance functions to be consistent respectively, which can be estimated by any machine learning methods, including parametric or semi-parametric methods.
Also, $\hat \tau_{reg}(x)$ has the same consistency result as $\hat \tau_{naive}(x)$, both requiring the nuisance functions $\hat \mu_S^E(a,x)$, $\hat \mu_S^O(a,x)$, and $\hat \mu_Y^O(a,x)$ to be consistent. 
In Section \ref{sec: converg rate} we show they also share similar asymptotic properties theoretically. 
This is reasonable since the design of the two-stage estimator $\hat \tau_{reg}(x)$ is motivated by $\hat \tau_{naive}(x)$ as well as the identification result in Eq. \eqref{eq: identi} in Theorem \ref{theo: identifi}.
Additionally, Lemma \ref{lem: baselines consis} can be seen as the generalization of two-stage estimators in traditional causal inference \cite{curth2021nonparametric}, which do not consider long-term effect estimation and also do not consider the data combination scenarios. 
In our paper, the considered estimators above are much more different and complex than the ones in \cite{curth2021nonparametric} 
and further, to achieve consistency, our estimators require more nuisance functions to be consistent,
since in our setting, the causal graphs in Fig. \ref{fig: causal graph}, the defined nuisance functions in Eq. \eqref{defined nuisance}, and the identification result in Eq. \eqref{eq: identi} are much different.

\subsection{Multiple Robust Estimator}
\label{subsec: mr estimaotr process}

As shown in Lemma \ref{lem: baselines consis}, the regression-based estimator $\hat \tau_{reg}(x)$ and propensity-based estimator $\hat \tau_{pro} (x)$ are consistent only if their nuisance functions are consistently estimated.
However, this assumption is easily violated when misspecified parametric regression methods are used to estimate the required nuisance functions.
In contrast, multiple robust estimators,  which incorporate several nuisance functions, still yield consistent estimates of effects as long as part of the nuisance functions are consistent. 
Beyond giving more chances at consistent estimations of nuisance functions, multiple robust estimators can also attain faster rates of convergence than their nuisance functions when all of the nuisances are consistently estimated. 
This advantage is particularly significant when employing flexible machine learning models with universal approximation properties, such as neural networks.

To this end, we design the multiple robust estimator, denoted as $\hat \tau_{mr}(x)$. Specifically, the estimator can be constructed by:
\begin{itemize}
    \item [S1.] Fitting nuisance functions $\hat \pi^E(a,x)$, $\hat \pi^O(a,x)$, $\hat \pi^G(a,x)$, $\hat \pi^E(a,x)$, $\hat \pi^O(a,x)$, and $\hat \pi^G(a,x)$;
    \item [S2.] Regressing the pseudo outcome $\hat Y_{mr}$ on covariates $X$ to obtain $\hat \tau_{mr} (x)$, i.e., $\hat \tau_{mr} (x)=\hat \bE [\hat Y_{mr}|X=x]$, where the pseudo outcome $\hat Y_{mr}$ follows
    \begin{equation} 
    \begin{aligned}        
    & \hat Y_{mr} \\
    = & \frac{(-1)^{1-A}}{1-A+(-1)^{1-A} \hat \pi ^E(X)} \frac{\mathbb I(G=E)}{p(G=O)}\\
    & \times
    (S-\hat \mu_S^E(A,X) )(\frac{1}{\hat \pi ^G(X)}-1) \\
    & + \frac{\mathbb I(G=O)}{p(G=O)}\frac{(-1)^{1-A}}{1-A+(-1)^{1-A} \hat \pi ^O(X)}\\
    & \times (Y- \hat\mu_Y^O(A,X) -S + \hat\mu_S^O(A,X))  \\
    & + \hat \mu_Y^O(1,X) - \hat \mu_Y^O(0,X) + \hat \mu_S^E(1,X)  \\
     & - \hat \mu_S^E(0,X) + \hat \mu_S^O(0,X) - \hat \mu_S^O(1,X)  ;
    \end{aligned}
    \end{equation}
\end{itemize}

The above estimator shares a multiple robustness property, as shown in the following lemma.

\begin{lemma} [MR Estimator Consistency] \label{lem: mr consis}
    Suppose Assumptions \ref{assum: consist}, \ref{assum: positi}, \ref{assum: internal validity of obs}, \ref{assum: internal validity of exp}, \ref{assum: external validity of exp} and \ref{assum: equ bias} hold, then $\hat \tau_{mr}(x)$ is consistent as long as one of the following sets of nuisance functions is consistent:
    \begin{equation} 
      \begin{aligned}
            & \{ \hat \mu^O_S(a,x), \hat \mu^E_S(a,x),\hat \mu^O_Y(a,x) \} ; \\
            & \{ \hat \pi^E(x), \hat \pi^O(x),  \hat \pi^G(x) \}; \\
            & \{ \hat \mu^E_S(a,x),  \hat \pi^O(x) \}; \\
            & \{ \hat \pi^E(x), \hat \mu^O_S(a,x), \hat \mu^O_Y(a,x), \hat \pi^G(x) \}.
      \end{aligned}
    \end{equation}
\end{lemma}

Proof can be found in Appendix \ref{app: proof of mr consi}.
Similarly to regression-based and propensity-based estimators, the nuisance functions above can be estimated by any machine-learning method, including parametric or semi-parametric methods. 
Compared with the consistency result in Lemma \ref{lem: baselines consis}, multiple robust estimator poses much weaker requirements on the consistent estimation of the nuisance functions.
Note that, the first set of the nuisance functions is exactly the same as that required by the regression-based estimators $\hat \tau_{reg}(x)$, and the second set is exactly the same as that required by the propensity-based estimator $\hat \tau_{pro}(x)$. 
The multiple robust estimator $\hat \tau{mr}(x)$ remains consistent even when both $\hat \tau_{reg}(x)$ and $\hat \tau_{pro}(x)$ are inconsistent, provided that either the third or fourth set is consistent. 
In the next section, we provide in-depth theoretical analyses and show the multiple robust estimator has better asymptotic properties over regression-based and propensity-based estimators.

\section{Theoretical Analysis} \label{sec: converg rate}

In this section, we theoretically analyze the proposed baseline estimators and the multiple robust estimator. 
Specifically, 
we compare different estimators for the long-term heterogeneous causal effects in asymptotic and finite sample settings.
The theoretical analysis provides insights and guides principled choices between different proposed estimators. 

Throughout, we denote stochastic boundedness with $O_p$.
Let $a \lesssim b$ denote the relation $a \leq Cb$ for some universal constant $C$, and let $a \asymp b$ denote both $a/b$ and $b/a$ are bounded. In order to compare the performances of different estimators, it is useful to analyze under what conditions the estimators can behave like the oracle estimator that can regress $(Y(1)-Y(0))$ on $X$ directly.

\begin{definition}[Oracle rate] \label{def: oracle rate}
Let $\tilde \tau(x) = \hat {\mathbb E} _ n [Y(1)-Y(0)|X=x]$ denote an oracle (infeasible) estimator that directly regresses the difference $(Y(1)-Y(0))$ on $X$, and let $R^*(n)$ be its error under some loss, e.g, the oracle mean squared error is $\mathbb E[(\tilde \tau (x)-\tau(x))^2]$. We refer to  $R^*(n)$ as the oracle rate.
\end{definition}

At various points, we refer to $s$-smooth functions contained in the H\"older ball $\mathcal H _d(s)$, associated with the minimax rate \cite{stone1980optimal} of $n^{\frac{1}{2+d/s}}$ where $d$ is the dimension of $\mathcal X$. Formally, we give the following definition.

\begin{definition}[H\"older ball]  \label{def: holder ball}
The H\"older ball $\mathcal H _d(s)$ is the set of $s$-smooth functions $f: \mathbb R^d \rightarrow \mathbb R$ supported on $\mathcal X \subseteq \mathbb R^d$ that are $\lfloor s \rfloor $-times continuously differentiable with their multivariate partial derivatives up to order $\lfloor s \rfloor $ bounded, and for which
\begin{equation*}
    | \frac{\partial ^m f }{\partial^{m_1} \cdots \partial^{m_d} }(x) - \frac{\partial ^m f }{\partial^{m_1} \cdots \partial^{m_d} }(x^\prime) | \lesssim \| x - x^\prime\|^{s- \lfloor s \rfloor }_2,
\end{equation*}
$\forall x, x^\prime$ and $m=(m1, \cdots, m_d)$ such that $\Sigma_{j=1}^d m_j = \lfloor s \rfloor$. 
\end{definition}

To derive the asymptotic bound on the convergence rate of our estimators, we make the following smoothness and boundedness assumptions.

\begin{assumption}[Smoothness Assumption] \label{asmp:smooth}
    We assume that the HLCE and the nuisance functions satisfy: (1)  the HLCE $\tau$ is $\kappa$-smooth;  (2) $\pi^E$, $\pi^O$, $\pi^G$, $\mu_S^E$, $\mu_S^O$, and $\mu_Y^O$ are $\alpha$-smooth, $\beta$-smooth, $\gamma$-smooth, $\eta$-smooth,  $\delta$-smooth, and $\zeta$-smooth, respectively.
\end{assumption}

\begin{assumption}[Boundedness Assumption] \label{asmp:bounded}
    We assume that the following nuisance functions and estimates are bounded, i.e., for some $\epsilon^E, \tilde \epsilon^E, \epsilon^O, \tilde \epsilon^O, \epsilon^G, \tilde \epsilon^G >0$, we have 
    $ \epsilon^E <  \pi^E(x) < 1- \epsilon^E$,
    $\tilde \epsilon^E < \hat \pi^E(x) < 1-\tilde \epsilon^E$,
    $ \epsilon^O <  \pi^O(x) < 1- \epsilon^O$,  
    $\tilde \epsilon^O < \hat \pi^O(x) < 1-\tilde \epsilon^O$,  
    $\epsilon^G < \pi^G(x) < 1-\epsilon^G$,
    and $\tilde \epsilon^G < \hat \pi^G(x) < 1-\tilde \epsilon^G$.
    We also assume that the following nuisance functions are bounded, i.e., 
    for $C^E, C^O, C^O_Y >0$, we have 
    $ | \mu ^E_S(a,x) | <C^E$, $ | \mu ^O_S(a,x) | <C^O$, and $ | \mu ^O_Y(a,x) | <C^O_Y$.
\end{assumption}

The Assumptions \ref{asmp:smooth} and \ref{asmp:bounded} are commonly used and in line with previous works on theoretical analyses of such two-stage heterogeneous effect estimators in different settings\cite{kennedy2023towards, curth2021nonparametric,frauen2023estimating}.
Specifically, Assumption \ref{asmp:smooth} quantifies the difficulty of nonparametric regression of nuisance functions, allowing us to systematically compare the performances between different HLCE estimators.
This assumption can also be replaced with a sparsity assumption on the nuisance functions when data is high-dimensional (See Appendix \ref{app: sparse}).
Assumption \ref{asmp:bounded} is standard, ensuring that both some of the nuisance functions and their estimates are bounded. 
Violations of Assumption \ref{asmp:bounded} may occur when the covariate distributions between different groups are extremely imbalanced, e.g., $\epsilon^G < \pi^G(x) < 1-\epsilon^G $ can be violated when almost no sample are available in the experimental data. 
However, in many real-world applications, experiments are often artificially designed so that this assumption can hold.

We now state our main theoretical results: the upper bounds on the oracle rate $R^*(n)$ of our proposed estimators. To obtain our bounds, we leverage the same sample splitting technique from \cite{kennedy2023towards}, which randomly splits the datasets into two independent sets and applies them in the regressions of the first step and second step respectively. Such a technique is originally used to analyze the convergence rate of the double robust conditional average treatment effect estimation in the traditional setting \cite{kennedy2023towards} and later is adapted to several other methods \cite{curth2021nonparametric, frauenestimating}, yet \textbf{not} for the HLCE estimation.

\begin{theorem} [Convergence Rate] \label{thm: convergence rate}
Suppose the first and second training steps of $ \hat\tau_{mr} (x)$ are train on two independent datasets of size $n$ respectively, and suppose  Assumptions \ref{assum: consist}, \ref{assum: positi}, \ref{assum: internal validity of obs}, \ref{assum: internal validity of exp}, \ref{assum: external validity of exp}, \ref{assum: equ bias} and \ref{asmp:bounded} hold, then we have 
\begin{equation}
  \begin{aligned}
     &  \hat\tau_{mr} (x)  -\tau(x)  = O_p( R^*(n) + \\
     &  (r_{\pi^E}(n)+r_{\pi^G}(n)) r_{\mu_S^E}(n) +  r_{\pi^O}(n) (r_{\mu_Y^O}(n) + r_{\mu_S^O}(n)) ),
  \end{aligned}
\end{equation}
where $r_{\circ}(n)$ denotes the risk of nuisance function $\circ$, e.g., $r_{\pi^E}(n)$ correspondingly to $\pi^E(x)$.
\end{theorem}

Proof can be found in Appendix \ref{app: proof of mr rate}. Again, Theorem \ref{thm: convergence rate} shows the multiple robustness of the estimator $\hat \tau_{mr}(x)$, since the last two terms are product terms. 
For a product term to be consistent, we only require one factor to be consistent, e.g., $(r_{\pi^E}(n)+r_{\pi^G}(n)) r_{\mu_S^E}(n)=o(1)$ holds as long as $r_{\pi^E}(n)+r_{\pi^G}(n)=o(1)$ or $ r_{\mu_S^E}(n)=o(1)$ hold.
More attractively, if each term is consistent, the estimator enjoys a faster convergence rate than non-multiple robust estimators whose convergence rate generally matches that of the nuisance function estimate.
Moreover, if the experiment design is already known (i.e., $\pi^E(x)$ and $\pi^G(x)$ are known), the estimator $\hat \tau_{mr}(x)$ becomes consistent if either $\hat \pi^O(x)$ is consistent or $\hat \mu^O_Y(x)$ and $\hat \mu_S^O(x)$ are consistent.
Based on Theorem \ref{thm: convergence rate} and Assumption \ref{asmp:smooth}, we further analyze the asymptotic properties of the baseline estimators and the multiple robust estimator in the following theorem and corollary, 
which provides comparisons between convergence rates of different estimators, thus guiding principled choices between these estimators.

\begin{theorem} [MR Estimator Convergence Rate] \label{theo: mf convergence rate}
    Suppose assumptions in Theorem \ref{thm: convergence rate} and Assumption \ref{asmp:smooth} hold, then we have
    \begin{equation}
    \begin{aligned}
        & \hat\tau _{mr} (x)-\tau(x) \\
        = &  O_p(n^{-\frac{1}{2+d/\kappa}} 
         + n^{-(\frac{1}{2+d/\alpha}+\frac{1}{2+d/\eta})} 
         + n^{-(\frac{1}{2+d/\gamma}+\frac{1}{2+d/\eta})} 
         \\  &
         + n^{-(\frac{1}{2+d/\zeta}+\frac{1}{2+d/\beta})} 
         +n^{-(\frac{1}{2+d/\delta}+\frac{1}{2+d/\beta})} ).
    \end{aligned}
    \end{equation}
   And the MR estimator is oracle efficient if
    \begin{equation}
        \begin{aligned}
       \frac{1}{\kappa} \geq 
             \max \{ &
             \frac{d^2-4\alpha\eta}{4d\alpha\eta + d^2\alpha+d^2\eta},
             \frac{d^2-4\gamma\eta}{4d\gamma\eta + d^2\gamma+d^2\eta},
             \\ &
             \frac{d^2-4\zeta\beta}{4d\zeta\beta + d^2\zeta+d^2\beta},
             \frac{d^2-4\delta\beta}{4d\delta\beta + d^2\delta+d^2\beta}
             \} .
        \end{aligned}
    \end{equation}
\end{theorem}

\begin{corollary} [Baseline Estimators Convergence Rate]
\label{coro: reg conv rate}
    Suppose the first and second training steps of $ \hat\tau_{reg} (x)$ and $ \hat\tau_{pro} (x)$ are train on two independent datasets of size $n$ respectively, and suppose Assumptions \ref{assum: consist}, \ref{assum: positi}, \ref{assum: internal validity of obs}, \ref{assum: internal validity of exp}, \ref{assum: external validity of exp}, \ref{assum: equ bias}, \ref{asmp:smooth}, and \ref{asmp:bounded} hold.
    For the estimators $\hat{\tau}_{naive} (x) $ we have
    \begin{equation} 
      \begin{aligned}
       & \hat\tau_{naive} (x) -\tau(x) \\
       = & O_p( n^{-\frac{1}{2+d/\eta}} + n^{-\frac{1}{2+d/\delta}} + n^{-\frac{1}{2+d/\zeta}}  ).
      \end{aligned}
    \end{equation}
    For the estimators $ \hat\tau_{reg} (x) $ we have
    \begin{equation} 
      \begin{aligned}
        & \hat\tau_{reg} (x) -\tau(x)  \\
        = &  O_p(n^{-\frac{1}{2+d/\kappa}}+ n^{-\frac{1}{2+d/\eta}} + n^{-\frac{1}{2+d/\delta}} + n^{-\frac{1}{2+d/\zeta}}  ),
      \end{aligned}
    \end{equation}
    and the estimator $\hat\tau_{reg} (x)$ is oracle efficient if
       \begin{equation} 
      \begin{aligned}
        \kappa \leq \min\{\eta, \delta, \zeta \}.
      \end{aligned}
    \end{equation}
    For the estimators $\hat\tau_{pro} (x) $ we have
        \begin{equation} 
      \begin{aligned}
        & \hat\tau_{pro} (x) -\tau(x) \\
        = & O_p(n^{-\frac{1}{2+d/\kappa}}+ n^{-\frac{1}{2+d/\alpha}} + n^{-\frac{1}{2+d/\beta}} + n^{-\frac{1}{2+d/\gamma}} ),
      \end{aligned}
    \end{equation}
    and the estimator $\hat\tau_{pro} (x)$ is oracle efficient if
   \begin{equation} 
      \begin{aligned}
        \kappa \leq \min\{\alpha, \beta, \gamma \}.
      \end{aligned}
    \end{equation}
\end{corollary}

Proof can be found in Appendix \ref{app: proof of mr rate under smooth} and \ref{app: proof of baseline rate}.
In practice, it is commonly assumed that the heterogeneous causal effect $\tau(x)$ is smoother than the nuisance functions, i.e., $\kappa > \max\{\alpha, \gamma,\eta, \delta, \zeta, \beta\}$.
In this sense, for the estimators $\hat\tau_{reg} (x)$ and $\hat\tau_{pro} (x)$, they are unlikely to attain the oracle rate.
And asymptotically, $\hat\tau_{reg} (x)$ and $\hat\tau_{naive} (x)$ will attain the same rate, thus we expect $\hat\tau_{reg} (x)$ and $\hat\tau_{naive} (x)$ achieve similar performance.
Compared the MR estimator $\hat\tau_{mr} (x)$ with these baselines estimators, we prefer $\hat \tau (x)$ since it achieves a faster rate than that of these estimators.
Moreover, the MR estimator $\hat\tau_{mr} (x)$ is easier to attain the oracle rate as its rate contains several product terms.
In the case where the heterogeneous effect $\tau(x)$ is of similar smoothness as the nuisance functions, these four estimators are then expected to perform similarly. 
Instead of assuming the smoothness condition of the nuisance functions, similar analyses can be performed by relying on different assumptions on the problem structure. 
In Appendix \ref{app: sparse}, we consider the sparsity assumption instead of Assumption \ref{asmp:smooth}, leading to analogous conclusions in terms of the relative performance of the different estimators.

\section{Neural Network-based Estimator} \label{sec: NN est}

\begin{figure}[!h]
    \centering
    \includegraphics[width=1.\linewidth]{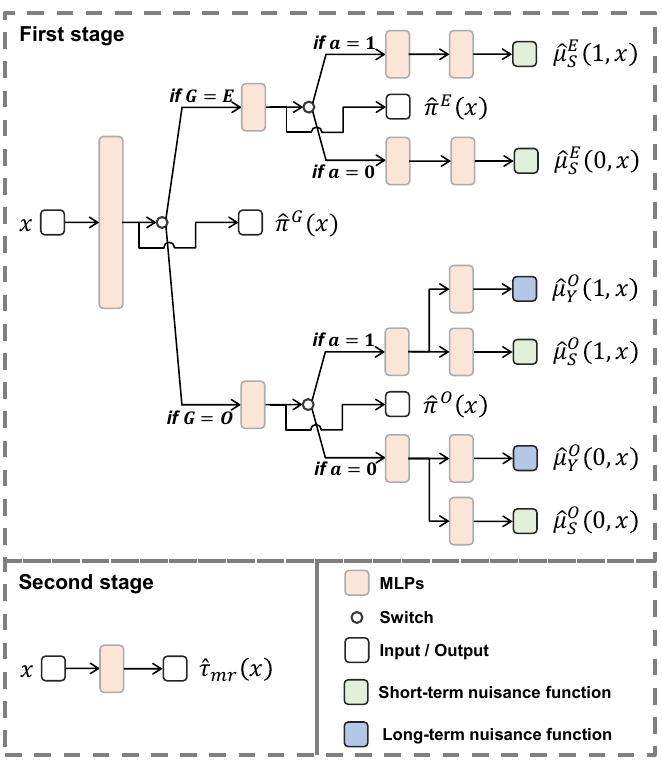}
    \caption{Our neural network-based model architecture of the MR estimator $\hat \tau_{mr}(x)$. Pink blocks denote MLPs. White blocks denote inputs or outputs. Green blocks denote short-term nuisance functions. Blue blocks denote long-term nuisance functions. White circles denote switches. Both learning stages are implemented using neural networks. 
    The top figure shows our first-stage learning, where we learn the shared representations across experimental and observational data, treated and control groups, as well as short- and long-term outcome predictions. 
    The bottom left figure illustrates our second-stage learning, where we regress the pseudo outcome $\hat Y_{mr}$ on the covariates $X$.}
    \label{fig: model}
\end{figure}

In previous sections, we provide a comprehensive theoretical analysis of the proposed long-term effect estimators in terms of their asymptotic properties. 
Note that, all of these estimators can be implemented by any off-the-shelf regression estimators. 
In this section, we provide a practical implementation of the multiple robust estimator $\hat \tau_{mr}(x)$ based on a tailored deep neural network
(we also provide the implementations of the regression-based estimator $\hat \tau_{reg}(x)$ and the propensity-based estimator $\hat \tau_{pro}(x)$ in a similar manner in Appendix \ref{app: baseline model}).
As shown in Figure \ref{fig: model}, our model consists of two separate Multi-Layer Perceptron (MLPs)-based estimation stages, which is consistent with the framework of the two-stage learning process in Section \ref{subsec: mr estimaotr process} and Lemma \ref{lem: mr consis}. 


In the first stage, inspired by Tarnet \cite{shalit2017estimating, johansson2022generalization}, we employ shared representations for estimators of all nuisance functions. 
Such a technique is widely applied in conditional causal effect estimators in different scenarios (e.g., \cite{louizos2017causal, chen2024doubly}), based on which, estimators have shown to be more efficient in finite sample regimes than those estimating nuisance functions using separate estimators (e.g., T-learner \cite{kunzel2019metalearners}). 
Hence, we propose to leverage shared representations between different groups. 
Unlike existing work that only shares representations between treated and control groups, we also employ shared representations between different data sources, i.e., experimental data and observational data, and between short and long-term outcomes. 
Specifically, as shown in Figure \ref{fig: model}, we first learn a shared representation between different data sources, which is also used to predict the nuisance function $\pi^G(x)$. 
Then, for experimental data $G=E$, we learn a shared representation between treated and control groups, which is used to output experimental nuisance functions $ \mu_S^E(1,x)$, $\mu_S^E(0,x)$ and $ \pi^E(x)$. 
Similarly, for observational data $G=O$, we learn a shared representation between treated and control groups, and we also learn a shared representation between short and long-term outcomes, which together output observational nuisance functions $\mu_Y^O(1,x)$, $\mu_Y^O(0,x)$, $\mu_S^O(1,x)$, $\mu_S^O(0,x)$ and $\pi^O(x)$.

In the second stage, we construct pseudo outcomes based on the first-stage output and perform the pseudo outcome regression using a simple MLP to obtain $\hat \tau_{mr}(x)$.

As for the baseline estimators, we employ similar shared representations to construct $\hat \tau_{reg}$ and $\hat \tau_{pro}$.
We further provide their model architectures in Appendix \ref{app: baseline model}.
Additionally, following existing two-stage methods built on neural networks \cite{curth2021nonparametric, frauenestimating}, we use all data for both regression stages, while our theoretical analyses rely on the sample splitting technique. Using all data for both stages has shown to perform better in practice especially when the models are implemented by neural networks.

\section{Experiments}  \label{sec: exp}

In this section, we conduct experiments to verify the effectiveness and correctness of our proposed methods. Specifically, we answer the following research questions (RQs):
\begin{itemize}
    \item \textbf{RQ1 (Multiple Robustness):} Can $\hat \tau_{mr}(x)$ achieve multiple robustness?
    \item \textbf{RQ2 (Accuracy):} Can $\hat \tau_{mr}(x)$ and baselines $\hat \tau_{reg}(x)$ and $\hat \tau_{pro}(x)$ achieve accurate long-term effect estimation?
    \item \textbf{RQ3 (Sample Sensitivity):} Are our proposed methods sensitive to sample size?
    \item \textbf{RQ4 (Comparision Performance):} Can $\hat \tau_{mr}(x)$ outperform other methods in terms of long-term effect estimation?

\end{itemize}

\begin{figure*}[!t]
    \centering
    \subfloat[Fix $n_o=2000$, vary $n_e$]{\includegraphics[width=.47\textwidth]{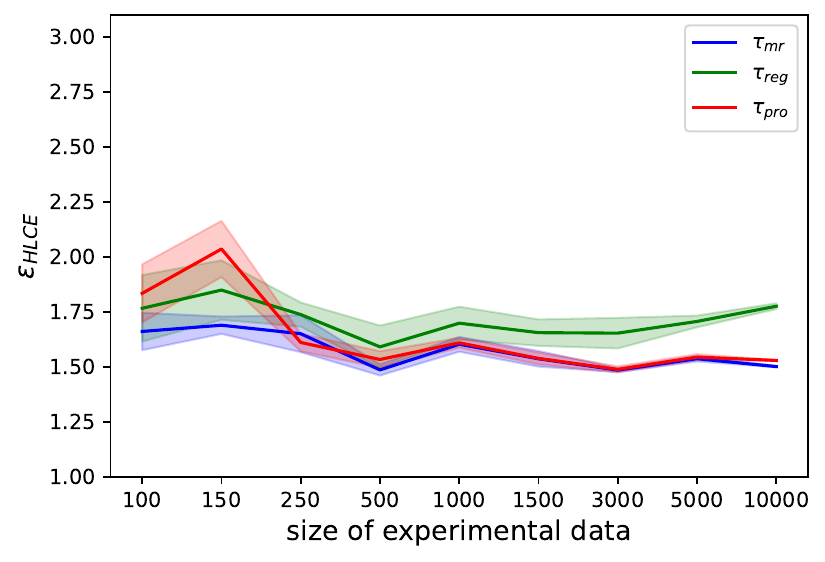}
    \label{fig: ite, fix o vary e}}
    \hspace{18pt}
    \subfloat[Fix $n_o=2000$, vary $n_e$]{\includegraphics[width=.47\textwidth]{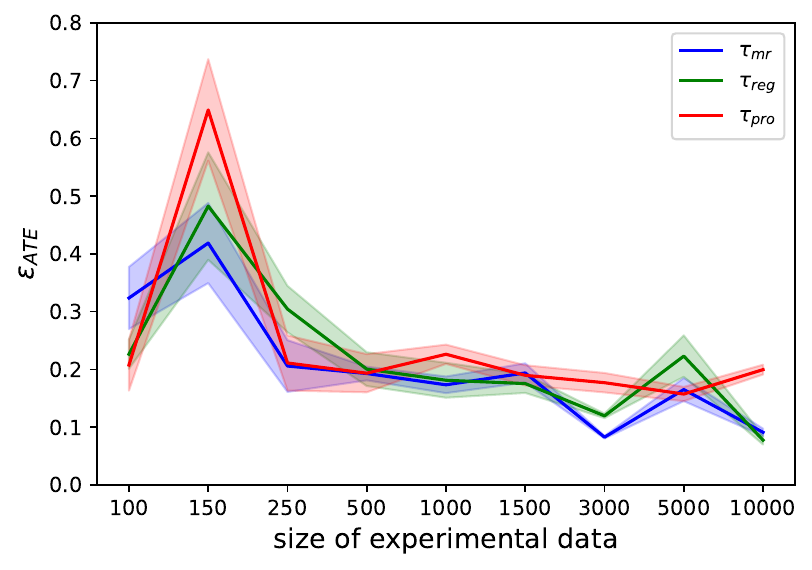}
    \label{fig: ate, fix o vary e}}
    \\
    \subfloat[Fix $n_e=1000$, vary $n_o$]{\includegraphics[width=.47\textwidth]{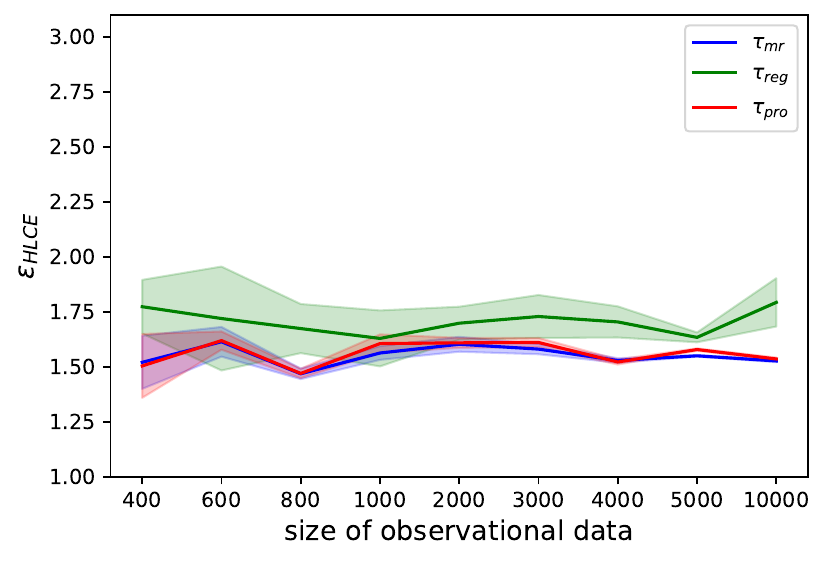}
    \label{fig: ite, fix e vary o}}
    \hspace{18pt}
    \subfloat[Fix $n_e=1000$, vary $n_o$]{\includegraphics[width=.47\textwidth]{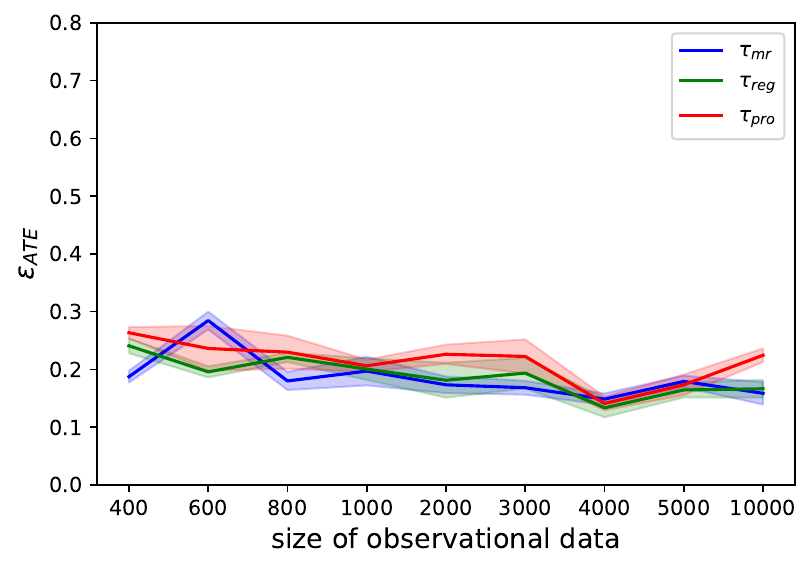}
    \label{fig: ate, fix e vary o}}
    \caption{Results on dataset 2. 
    Fig.\ref{fig: ite, fix o vary e} reports the PEHE of the heterogeneous effect estimation with a fixed size of experimental data and a varying size of observational data. 
    Fig.\ref{fig: ate, fix o vary e} reports the absolute error of the average effect estimation with a fixed size of experimental data and a varying size of observational data. 
    Fig.\ref{fig: ite, fix o vary e} reports the PEHE of the heterogeneous effect estimation with a fixed size of observational data and a varying size of experimental data. 
    Fig.\ref{fig: ite, fix o vary e} reports the absolute error of the average effect estimation with a fixed size of observational data and a varying size of experimental data.}
    \label{fig: simulated dataset 2}
\end{figure*}

\subsection{Experimental Set up}

\subsubsection{Datasets}

To answer the research questions above, we conduct extensive experiments on two semi-synthetic datasets and multiple synthetic datasets.
For these datasets, we randomly split them into train/validation/test splits with ratios 63/27/10.

As for the synthetic datasets, we generate two different datasets to answer RQ1, RQ2, and RQ3. 
First of all, to answer RQ1, our data generation process follows \cite{kallus2018removing}, such that we can obtain specific forms of all nuisance functions, in order to verify the multiple robustness property. The size of experimental data $n_e$ and observational data $n_o$ is $10000$ and $15000$ respectively.
This dataset is denoted as \textbf{Dataset 1}.
Secondly, to answer RQ2 and RQ3, our data generation process partly follows \cite{frauen2023estimating}, where each nuisance function is simulated from Gaussian processes using the prior induced by the Matern kernel \cite{Ras2005GPfML}, which can control the smoothness of nuisance functions. This dataset is denoted as \textbf{Dataset 2}. In this dataset, we vary the sample size to better answer RQ2 and RQ3, i.e., the size of experimental data $n_e$ satisfying $ n_e \in \{ 100,150,250,500,\bold{1000},1500,3000,5000,10000 \}$ and the size of observational data $n_o$ satisfying $n_o \in \{ 400,600,800,1000,\bold{2000},3000, 4000,5000,10000 \}$ where bold numbers are the default values. 
The detailed steps to generate the datasets 1 and 2 are given in Appendix \ref{app: additional exp}.

As for the semi-synthetic datasets, following the existing work on long-term causal inference \cite{cheng2021long, cai2024long, yang2024estimating}, experiments are conducted on two widely used dataset, the Infant Health and Development Program (\textbf{IHDP}) dataset \cite{hill2011bayesian} and the \textbf{News} dataset \cite{johansson2016learning}. The IHDP dataset is collected from a real-world randomized controlled experiment, which aims to evaluate the effect of
high-quality child care and home visits on the children’s cognitive test scores, and the News dataset was originally introduced by \cite{johansson2016learning} to simulate the opinions of a media consumer exposed to multiple news items based on the NY Times corpus \cite{newman2008bag}. Specifically, we reuse the covariates in these datasets, divide covariates into observed $X$ and unobserved $Z$, and then simulate group indicator $G$, treatment $A$, short-term outcome $S$ and long-term outcome $Y$ following the causal graphs in Figure \ref{fig: causal graph} and Assumptions \ref{assum: consist}, \ref{assum: positi}, \ref{assum: internal validity of obs}, \ref{assum: internal validity of exp}, \ref{assum: external validity of exp} and \ref{assum: equ bias}.
Complete details are given in Appendix \ref{app: additional exp}.

\subsubsection{Baselines}

We compare our designed methods, denoted as \textbf{Ours ($\tau_{pro}(x)$)}, \textbf{Ours ($\tau_{reg}(x)$)}, and \textbf{Ours ($\tau_{mr}(x)$)} respectively, with  several baselines including a state-of-the-art neural network-based model and some statistical models. 
\begin{itemize}
    \item \textbf{LTEE}\cite{cheng2021long}: LTEE proposes an HTCE estimator under the unconfoundedness assumption, which minimizes the factual loss in terms of short-term and long-term outcome plus an extra IPM term that balances the representation between treated and control groups.
    \item \textbf{Athey et~al.} \cite{athey2020combining}: Athey et~al. propose a method to estimate long-term average causal effects under the latent unconfoundedness assumption, by imputing the missing long-term outcomes of observational data using a regression obtained by experimental data.
    \item \textbf{Ghassami et~al.} \cite{ghassami2022combining}: Ghassami et~al. propose to estimate long-term average effects under Assumption \ref{assum: equ bias}, based on the efficient influence function for average effects (Theorem 14 and Section 7.1.1 in \cite{ghassami2022combining}).
    \item \textbf{Naive ($\tau_{naive}(x)$)}: Naive is the estimator based on the identification result in Theorem \ref{theo: identifi}, consisting of four conditional regression models to estimate HDRC as shown in Eq. \eqref{eq: identi}. 
    \item \textbf{Ours ($\tau_{pro}(x), \tau_{reg}(x), \tau_{mr}(x)$)}: Our proposed estimators are constructed following the algorithms described in Sec. \ref{sec: estimator}.
\end{itemize}

\begin{table}[!h]
\renewcommand{\arraystretch}{1.1}
\caption{Hyper-parameter Space}     
\label{tab: hyper-param}
\centering 
\resizebox{\linewidth}{!}{
\begin{tabular}{c|c}
        \hline
        \textbf{hyper-parameter} & \textbf{space} \\ \hline
        learning rate & $\{1e-3, 1e-4\}$\\         
        weight decay & $\{1e-2, 1e-3, 1e-4\}$\\ 
        number of layers & $\{2,3\}$ \\ 
        number of hidden units & $\{16, 32, 64, 128\}$  \\ 
        batch size & $\{32, 64, 128\}$ \\ 
        dropout rate & $\{0, 0.05, 0.1, 0.2\}$ \\ \hline
    \end{tabular}
    }
\end{table}

\subsubsection{Implementation}

For a fair comparison, we implement baselines and our methods using MLPs, with the same hyper-parameter selection strategy. 
The hyper-parameter space is shown in Table \ref{tab: hyper-param}. 
Specifically, for LTEE, we use the official code available at \url{https://github.com/GitHubLuCheng/LTEE}.
For Athey et~al., Ghassami et~al., and naive, we use Tarnet-like model architecture for their nuisance functions, i.e., shared representation-based neural networks. 
Except for LTEE, we implement all methods using the PyTorch library \cite{shen2020pytorch}. 
All experiments are run on the NVIDIA GeForce RTX 2080 Ti and the NVIDIA Tesla K80.
Our code will be available upon acceptance.

\subsubsection{Metrics}

As for HLCE estimation, we report Precision in the Estimation of Heterogeneous Effect (PEHE) $\epsilon_{HLCE}= \sqrt{\frac{1}{n} \Sigma_{i=1}^{n} (\tau(x_i) - \hat \tau(x_i))^2}$. As for average long-term causal effect estimation, we report the absolute error $\epsilon_{ATE}= | \Sigma_{i=1}^{n} \tau(x_i) -\Sigma_{i=1}^{n}  \hat \tau(x_i)) |$. 
For all metrics, we report the mean values and deviations on the testsets by 10 times running. 
Note that, Athey et~al. and Ghassami et~al. are designed to estimate the long-term average effects, thus we only report $\epsilon_{ATE}$ for their methods.

\begin{table*}[!t]
\renewcommand{\arraystretch}{1.5}
\caption{Estimation errors regarding ATE and HLCE on the News and IHPD datasets. We report mean$(\pm$std$)$ results. Here / means these methods are not applicable since they are designed for average effect estimation. The best is \textbf{bolded} and the follow-up is \underline{underlined}.}     
\label{tab: real world}
\centering 
\resizebox{\linewidth}{!}{
\begin{tabular}{@{}l|c c | c c |c c | c c @{}}
        \hline
         & \multicolumn{4}{c|}{ \textbf{News dataset}}  & \multicolumn{4}{c}{\textbf{IHDP dataset}}  
         \\ \hline
         & \multicolumn{2}{c|}{Within-Sample}  & \multicolumn{2}{c|}{Out-of-Sample} & \multicolumn{2}{c|}{Within-Sample}  & \multicolumn{2}{c}{Out-of-Sample}  
         \\ \hline
         & $\epsilon_{HLCE}$ &$\epsilon_{ATE}$ & $\epsilon_{HLCE}$ &  $\epsilon_{ATE}$ & $\epsilon_{HLCE}$ &$\epsilon_{ATE}$ & $\epsilon_{HLCE}$ &  $\epsilon_{ATE}$
         \\ \hline
        \multirow{2}{*}{LTEE \cite{cheng2021long}}  &
        $ 11.7326  $  &  
        $ 26.2827  $ & 
        $ 11.9263  $ & 
        $ 11.3284  $ &
        $ 25.733  $ &
        $ \textbf{0.2305} $&
        $ 6.7268$&
        $ \textbf{0.3065}$\\ 
        & 
        $ (\pm 1.8993) $  &  
        $ (\pm 39.2936) $ & 
        $ (\pm 2.0928) $ & 
        $ (\pm 1.9673)  $ &
        $ (\pm 37.717)  $ &
        $ (\pm 0.3364) $&
        $ (\pm 2.0879) $&
        $ (\pm 0.4076) $ \\
        \hline
        \multirow{2}{*}{Athey et~al. \cite{athey2020combining}} &
        / &
        $ 7.8658  $ &
        / &
        / &
        / &
        $ 1.6562 $ &
        / &
        / \\
         &
        / &
        $ (\pm 8.7038 ) $ &
        / &
        / &
        / &
        $( \pm 1.6107 ) $ &
        / &
        / \\ \hline
        \multirow{2}{*}{AmirEmad et~al. \cite{ghassami2022combining}} &
        / &
        $ 3.2637 $ &
        / &
        / &
        / &
        $ 1.0054$ &
        / &
        / \\
         &
        / &
        $( \pm 2.7627) $ &
        / &
        / &
        / &
        $ (\pm  1.1920) $ &
        / &
        /
        \\ \hline 
         \multirow{2}{*}{Naive ($\tau_{naive}(x)$)}&
        $ 8.9774 $ &
        $ \underline{2.0026}  $ &
        $ 9.8386 $ &
        $ \underline{2.1214} $ &
        $ 4.8517$ &
        $ 1.2594$ &
        $ 4.4144$ &
        $ 1.0434 $ \\
        &
        $ (\pm 1.2604)  $ &
        $ (\pm 1.5035)  $ &
        $ (\pm 1.9651) $ &
        $ (\pm 1.6396) $ &
        $ (\pm 3.4751) $ &
        $ (\pm 1.1269) $ &
        $ (\pm 2.9621) $ &
        $ (\pm 0.8985) $ 
        \\ \hline
        \multirow{2}{*}{Ours ($\tau_{pro}(x)$) }&
        $ 13.8816 $&
        $ 4.0324 $&
        $ 14.2672 $ &
        $ 4.0175 $ &
        $ 6.7680 $ &
        $ 1.9160$&
        $6.3748 $&
        $ 2.0737$\\&
        $ (\pm 6.0930) $&
        $ (\pm 3.2876) $&
        $ (\pm4.2121 )$ &
        $ (\pm 3.1447) $ &
        $ (\pm 4.7324)$ &
        $ (\pm 1.9406)$&
        $ (\pm 4.0530)  $&
        $ (\pm 2.1481)$
        \\ \hline
        \multirow{2}{*}{Ours ($\tau_{reg}(x)$) }&
        $ \underline{8.2511} $ & 
        $ 2.0216 $ &
        $ \underline{9.1206} $ & 
        $ 2.1823 $  &
        $ \underline{4.8161} $ &
        $ 1.1782$&
        $ \underline{4.3639} $ &
        $ 0.9399 $ \\ &
        $ (\pm 1.2011)$ & 
        $ (\pm 1.5526)$ &
        $ (\pm 1.9319)$ & 
        $ (\pm 1.6068) $  &
        $ (\pm3.4864 )$ &
        $ (\pm1.0833 )$&
        $ (\pm 2.9616)$ &
        $ (\pm 0.8296)$
        \\ \hline
        \multirow{2}{*}{ Ours ($\tau_{mr}(x)$)} & 
        $ \textbf{7.9275}$ & 
        $ \textbf{1.7815}$& 
        $ \textbf{8.1379}$& 
        $ \textbf{1.9073}  $  &
        $\textbf{4.8074 }$& 
        $ \underline{0.9955}  $ &
        $ \textbf{4.3552}$ & 
        $\underline{0.8265}$ \\
         & 
        $ (\pm 0.8634)$ & 
        $ (\pm 1.3971)$& 
        $  (\pm 1.1961)$& 
        $  (\pm 1.2284) $  &
        $ (\pm 2.7767) $& 
        $  (\pm 0.6775) $ &
        $ (\pm 2.3752)$ & 
        $ (\pm 0.4972)$
        \\ \hline
    \end{tabular}
    }
\end{table*}
\subsection{Result Analysis}

\begin{figure}[!h]
    \centering
    \includegraphics[width=1.\linewidth]{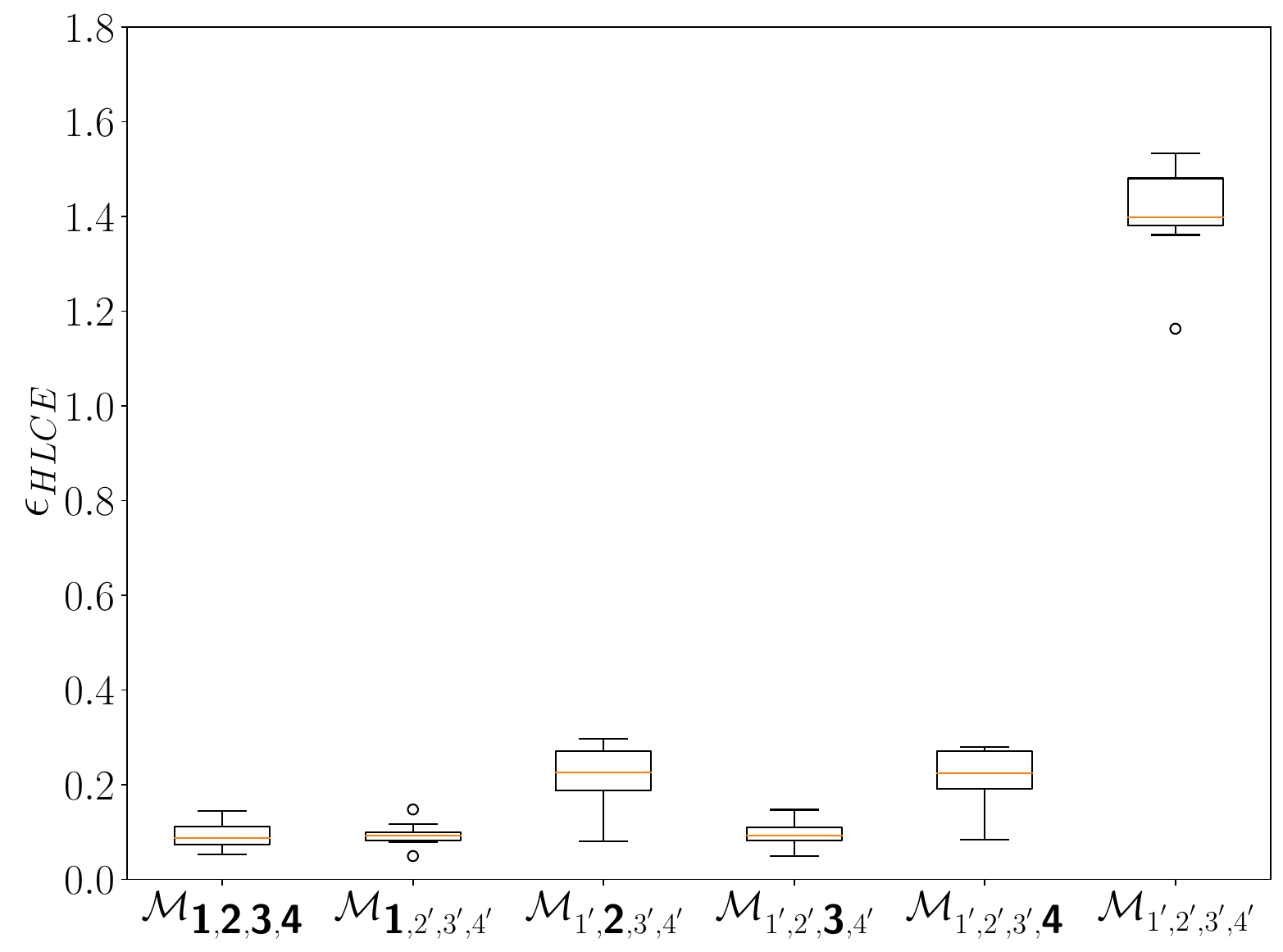}
    \caption{Results on dataset 1: Model Misspecification Experiments. The bold numbers mean the corresponding set of nuisance functions are correctly specified, and the numbers with $\prime$ mean the corresponding set of nuisance functions in Lemma \ref{lem: mr consis} are misspecified, e.g., $\mathcal{M}_{\textbf{1},\textbf{2},\textbf{3},\textbf{4}}$ means all sets of nuisance functions are correctly misspecified, 
    and $\mathcal{M}_{\textbf{1},{2^\prime},{3^\prime},{4^\prime}}$ means the first set of nuisance functions, i.e. $\{ \hat \mu^O_S(a,x), \hat \mu^E_S(a,x),\hat \mu^O_Y(a,x) \} $, is correctly specified and the rest of the sets are misspecified.
    }  
    \label{fig: misspecification}
\end{figure}

\subsubsection{RQ1: $\hat \tau _{mr}(x)$ achieves multiple robustness property} 

To verify whether $\hat \tau_{mr}(x)$ is multiple robust as shown in Lemma \ref{lem: mr consis}, we conduct an experiment on dataset 1, 
by implementing $\hat \tau _{mr}(x)$ using the (in)consistent parametric regressions as its nuisance functions (see Appendix \ref{app: parametric mr}).
The results are shown in Fig. \ref{fig: misspecification}.
As indicated by  Fig. \ref{fig: misspecification}, the models with at least one set of correctly specified nuisance functions can achieve a very low $\epsilon_{HLCE}$, while the model with all inconsistent nuisance functions performs poorly with approximated $\epsilon_{HLCE} \approx 1.5$.
This is reasonable since the multiple robustness property only requires at least one set of consistent nuisance functions and if all are misspecified, the result will be incorrect, leading to a high $\epsilon_{HLCE}$.
Overall, the result shown in Fig. \ref{fig: misspecification} demonstrates the correctness of Lemma \ref{lem: mr consis}, i.e., $\hat \tau _{mr}(x)$ achieves the multiple robustness property.

\subsubsection{RQ2: $\hat \tau _{mr}(x)$, $\hat \tau_{reg}(x)$, and $\hat \tau_{pro}(x)$ effectively estimate heterogeneous long-term effects} 

As shown in Fig. \ref{fig: simulated dataset 2}, we conduct experiments on dataset 2 to test the effectiveness of neural network-based $\hat \tau _{mr}(x)$, $\hat \tau_{reg}(x)$, and $\hat \tau_{pro}(x)$.
Overall, the three estimators achieve low $\epsilon_{HLCE}$ and $\epsilon_{ATE}$, which indicates the correctness of the proposed estimators.
In detail, regarding $\epsilon_{ATE}$, the three estimators achieve similar performances. 
Regarding $\epsilon_{HLCE}$, when compared with $\hat \tau_{reg}(x)$ and $\hat \tau_{pro}(x)$, $\hat \tau _{mr}(x)$ achieves slightly better performance with lower $\epsilon_{HLCE}$, especially when the size of data is small.
This is reasonable because  $\hat \tau _{mr}(x)$ shares the multiple robustness property and thus enjoys a faster convergence rate as shown in Theorem \ref{theo: mf convergence rate} and Corrollary \ref{coro: reg conv rate}, resulting in the higher efficiency with a limited data.

\subsubsection{RQ3:  $\hat \tau _{mr}(x)$, $\hat \tau_{reg}(x)$, and $\hat \tau_{pro}(x)$ work well across different sample sizes.} 

As shown in Fig. \ref{fig: simulated dataset 2}, with varying sample sizes, all estimators, i.e., $\hat \tau _{mr}(x)$, $\hat \tau_{reg}(x)$, and $\hat \tau_{pro}(x)$ perform stable and well with low $\epsilon_{HLCE}$ and $\epsilon_{ATE}$.
Specifically, as shown in Fig. \ref{fig: ite, fix o vary e} and \ref{fig: ate, fix o vary e}, with increasing sizes of experimental data, all estimators achieve better performances with smaller variances as expected.
Note that when the sample size is small, $\hat \tau _{mr}(x)$  performs best among them, and when the sample size is relatively large, $\hat \tau _{mr}(x)$  performs slightly better and more stably than $\hat \tau _{reg}(x)$ and $\hat \tau _{pro}(x)$.
This is reasonable because the multiple robust estimator is able to reduce biases by incorporating its multiple nuisance functions.
As shown in Fig. \ref{fig: ite, fix e vary o} and \ref{fig: ate, fix e vary o}, with increasing sizes of observational data, all estimators perform very well and stably, and similarly $\hat \tau _{mr}(x)$ consistently performs slightly better.
Overall, we conclude that the proposed estimators $\hat \tau _{mr}(x)$, $\hat \tau_{reg}(x)$, and $\hat \tau_{pro}(x)$ are not very sensitive to sample size, where $\hat \tau _{mr}(x)$ is the stablest one due to its multiple robustness property.

\subsubsection{RQ4: $\hat \tau _{mr}(x)$ can outperform all baselines in terms of effect estimation $\epsilon_{HLCE}$ and $\epsilon_{ATE}$} 
As shown in Tab. \ref{tab: real world}, we conduct experiments on two real-world datasets, IHDP and NEWS. 
Overall, the MR estimator $\tau_{mr}(x)$ performs the best as expected across different datasets. 
In detail, LTEE performs very unstably with large deviations since it cannot handle the unobserved confounders problem.
Athey et~al., based on LU assumption, can only partially address the unobserved confounders and thus result in biased estimations. 
The rest of the methods are all based on Assumption \ref{assum: equ bias} and can achieve unbiased estimation, resulting in a low  $\epsilon_{HLCE}$ and $\epsilon_{ATE}$.
$\tau_{naive}(x)$ and $\tau_{reg}(x)$ achieve very similar performance, which is consistent with our Corollary \ref{coro: reg conv rate}. $\tau_{pro}(x)$ exhibits unstable performance and this phenomenon also exists in the traditional setting \cite{curth2021nonparametric} and may be caused by the low signal-to-noise ratio and high variance in the associated pseudo outcome. 
$\tau_{mr}(x)$, as expected, performs consistently well, especially in heterogeneous effect estimation $\epsilon_{HLCE}$. This is due to its MR property, verifying Theorem \ref{theo: mf convergence rate} and Corrollary \ref{coro: reg conv rate} again. 

\section{Conclusion}
In this paper, we focus on the heterogeneous long-term causal effect estimation and propose several two-stage estimators, including the regression-based estimator, the propensity-based estimator, and the multiple robust estimator, which can be implemented using any off-the-shelf regression methods. 
We provide extensive theoretical analysis of the provided estimators, illustrating their asymptotical properties. 
We demonstrate that our multiple robust estimator is asymptotically optimal in theory among these estimators and enjoys an attractive multiple robustness property, which can effectively avoid the model misspecification problem in parametric regression and also lead to a faster convergence rate.
Practically, we design neural network-based architectures for the proposed estimators, which can learn the shared information between treated and control groups, as well as between observational and experimental data. 
Our extensive experiments across several synthetic and real-world datasets validate the effectiveness of the proposed estimators and the correctness of our theory.
The interesting next steps would be to explore different assumptions for effect identification and estimation and to explore architectures for more effective estimation, e.g., designing heterogeneous effect estimators with an additional proxy variable under proximal assumptions \cite{ghassami2022combining}.

\section*{Acknowledgments}
This research was supported in part by National Science and Technology Major Project (2021ZD0111501), National Science Fund for Excellent Young Scholars (62122022), Natural Science Foundation of China (U24A20233, 62206064, 62206061, 62476163) and CCF-DiDi GAIA Collaborative Research Funds (CCF-DiDi GAIA 202311). Weilin Chen's research was supported by the
China Scholarship Council (CSC).


{\appendices

\section{Additional Theoretical Results Under Sparsity Assumption}
\label{app: sparse}

In Section \ref{sec: converg rate}, we theoretically show the rate of our proposed estimators under the smoothness assumption (Assumption \ref{asmp:smooth}). In this section, instead of using the smoothness assumption, we make an assumption on the level of the sparsity of the nuisance functions and HLCE.

The sparsity assumption is often used in a high-dimensional setting where $n<d$ and $\mathcal X \in \mathbb R^d$.
This assumption is also in line with previous work on causal inference \cite{kennedy2023towards, frauenestimating, curth2021nonparametric}.
Following \cite{kennedy2023towards, frauenestimating, curth2021nonparametric}, we consider a class of functions with additive sparsity as defined in assumption M3 in \cite{yang2015minimax}. Specifically, a function $f$ satisfies additive sparsity if it depends on $d^\prime \asymp \min \{n^\nu, d\}$ variables for some $\nu \in (0,1)$ but admits an additive structure $f = \Sigma_{s=1}^k f_s$ where each component function $f_s$ depended on a small $d_s$ number of predictors.
A special case (i.e., M2 in \cite{yang2015minimax}) is the standard sparsity assumption that $f$ depends on a small subset of $d^\prime \leq \min \{n,d\}$. 
At the opposite extreme is another case where $f$ admits a completely additive structure ($d_s=1$ for all $s$).
To simplify our analysis, we assume all additive components $f_s$ have the same smoothness $p_s=p$, dimension $d_s=d^\prime$, and magnitude, and is linear in $\mathcal X$, thus its squared error of estimation, using the lasso estimator, can attain the minimax rate of $ \frac{d^\prime\log (d)}{n}$ (see \cite{bickel2009simultaneous} or Corollary 2 in \cite{raskutti2009lower}). Formally, we make the following sparsity assumption on our nuisance functions and HLCE.

\begin{assumption}[Sparsity Assumption] \label{asmp:sparse}
    We assume that all nuisance functions and HLCE are linear in $\mathcal X$ and satisfy: (1) the HLCE $\tau$ is $d_\tau$-sparse;  (2) $\pi^E$, $\pi^O$, $\pi^G$, $\mu_S^E$, $\mu_S^O$, and $\mu_Y^O$ is $d_{\pi^E}$-sparse, $d_{\pi^O}$-sparse, $d_{\pi^G}$-sparse, $d_{\mu_S^E}$-sparse, $d_{\mu_S^O}$-sparse, and $d_{\mu_Y^O}$-sparse, respectively.
\end{assumption}
Then we immediately conclude with the following theorem:

\begin{theorem} [Estimators Convergence Rate] \label{theo: rate under sparsity}
   Suppose the first and second training steps of our two-stage estimators are train on two independent datasets of size $n$ respectively, and suppose Assumptions \ref{assum: consist}, \ref{assum: positi}, \ref{assum: internal validity of obs}, \ref{assum: internal validity of exp}, \ref{assum: external validity of exp}, \ref{assum: equ bias}, \ref{asmp:bounded}, and \ref{asmp:sparse} hold, then we have
    \begin{equation} \small
    \begin{aligned}
          \hat\tau & _{naive} (x) -\tau(x) 
         =  O_p( \frac{d_{\mu_S^E} \log (d)}{n} 
         + \frac{d_{\mu_S^O} \log (d)}{n} 
         + \frac{d_{\mu_Y^O} \log (d)}{n}  ),
      \\
        \hat\tau & _{pro} (x) -\tau(x) 
        =  O_p( \frac{d_{\tau} \log (d)}{n} 
        + \frac{d_{\pi^E} \log (d)}{n} 
        + \frac{d_{\pi^O} \log (d)}{n} 
        \\  &
        + \frac{d_{\pi^G} \log (d)}{n}   ),
        \\
        \hat\tau & _{reg} (x) -\tau(x) 
        =  O_p( \frac{d_{\tau} \log (d)}{n}
        + \frac{d_{\mu_S^E} \log (d)}{n} 
        + \frac{d_{\mu_S^O} \log (d)}{n} 
        \\  &+ \frac{d_{\mu_Y^O} \log (d)}{n}  ), \\
         \hat\tau & _{mr} (x)-\tau(x) 
        =  O_p(  \frac{d_{\tau} \log (d)}{n}
         + \frac{(d_{\pi^E} + d_{\mu_S^E}) \log ^2(d)}{n^2} 
         \\  &
         + \frac{(d_{\pi^G} + d_{\mu_S^E}) \log ^2(d)}{n^2} 
         + \frac{(d_{\pi^O} + d_{\mu_Y^O}) \log ^2(d)}{n^2}
         \\  &
         + \frac{(d_{\pi^O} + d_{\mu_S^O}) \log ^2(d)}{n^2}  ).
    \end{aligned}
    \end{equation}
   And 
   the propensity-based estimator $\hat \tau_{pro}(x)$ is oracle efficient if
   $d_\tau \geq \max \{ d_{\pi^E}, d_{\pi^O}, d_{\pi^G}\}$, 
   the regression-based estimator $\hat \tau_{reg}(x)$ is oracle efficient if
   $d_\tau \geq \max \{ d_{\mu_S^E}, d_{\mu_S^O}, d_{\mu_Y^O} \}$,
   and 
   the MR estimator $\hat\tau _{mr} (x)$ is oracle efficient if 
   \begin{equation} 
    \begin{aligned}
    d_\tau \geq \max \{ &
    \frac{ (d_{\pi^E} + d_{\mu_S^E}) \log (d)}{n} , 
    \frac{ (d_{\pi^G} + d_{\mu_S^E}) \log (d)}{n} , \\&
    \frac{ (d_{\pi^O} + d_{\mu_Y^O}) \log (d)}{n} ,
    \frac{ (d_{\pi^O} + d_{\mu_S^O}) \log (d)}{n} \} .
        \end{aligned}
    \end{equation}
\end{theorem}

Proof can be found in Appendix \ref{app: proof of sparsity rate}.
We can draw analogous conclusions as presented in the main text under Assumption \ref{asmp:sparse}.
Generally, the HLCE is assumed to be simpler than its nuisance functions, i.e., 
$d_\tau \leq \min \{ d_{\mu_S^E}, d_{\mu_S^O}, d_{\mu_Y^O},d_{\pi^E}, d_{\pi^O}, d_{\pi^G} \}$ .
In this sense, $\hat \tau_{reg}(x)$ and $\hat \tau_{pro}(x)$ are hard to attain the oracle rate $\frac{d_{\tau} \log (d)}{n}$.
And also, $ \hat\tau_{naive} (x) $ and $\hat \tau_{reg}(x)$ are expected to perform similarly.
Additionally, since the rate of the MR estimator contains the product terms, its rate is faster than the baseline estimators $ \hat\tau_{naive} (x)$, $\hat \tau_{reg}(x)$, and $\hat \tau_{pro}(x)$, and it is also easier to attain the oracle rate.
We would thus prefer the MR estimator $\hat \tau_{mr}(x)$.

\section{Baseline Estimator Model Architecture} 
\label{app: baseline model}

The model architectures of baseline estimators $\hat \tau_{reg}(x)$ and  $\hat \tau_{pro}(x)$ are shown in Fig. \ref{fig: model reg} and Fig. \ref{fig: model pro} respectively. They share similar architecture with the multiple robust estimator $\hat \tau_{mr}(x)$, using the same shared representation learning technique.

\begin{figure}[!h]
    \centering
    \includegraphics[width=.95\linewidth]{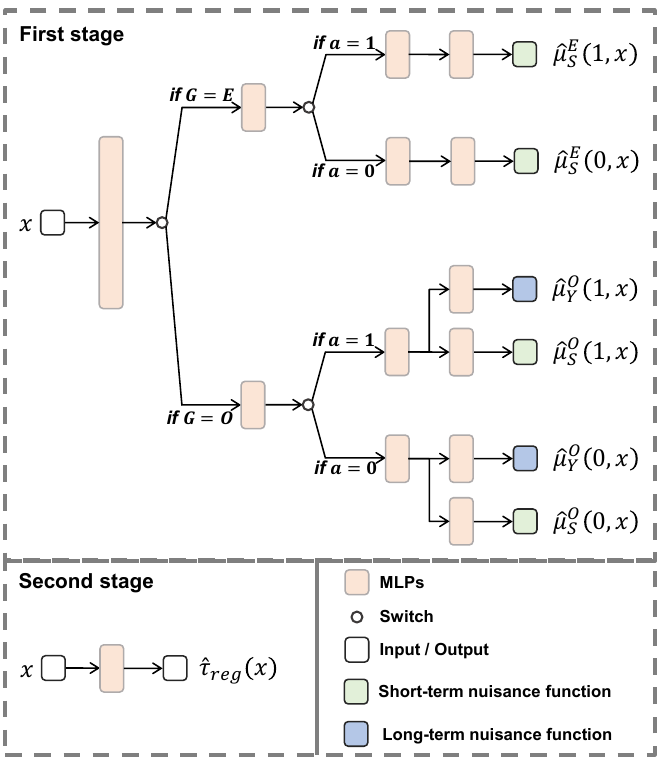}
    \caption{Our neural network-based model architecture of $\hat \tau_{reg}(x)$. Gray blocks denote MLPs.}
    \label{fig: model reg}
\end{figure}

\begin{figure}[!h]
    \centering
    \includegraphics[width=.95\linewidth]{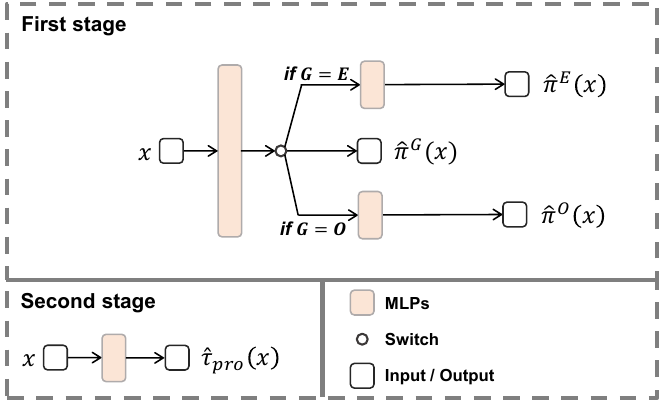}
    \caption{Our neural network-based model architecture of $\hat \tau_{pro}(x)$. Gray blocks denote MLPs.}
    \label{fig: model pro}
\end{figure}

\section{Experimental Details}
\label{app: additional exp}

\subsection{Data Generation Process}
\label{app: dgp}

\textbf{Dataset 1:} The data generation process is partly following \cite{kallus2018removing} such that we can obtain specific forms of all nuisance functions, in order to verify the MR property. Specifically, 
We first generate the treatments as follows: $A \sim \text{Bernoulli}(0.5)$. Then we generate the observed $X$ and the unobserved $U$ as follows:
   \begin{equation} \small
    \begin{aligned}
       & (X,U)|A,G=E \sim \mathcal N ([\frac{2A-1}{2},0], 
       \begin{bmatrix} 1 & 0 \\ 0 & 1
       \end{bmatrix})
       \\  
       & (X,U)|A,G=O \sim \mathcal N ([\frac{1-2A}{2},0], 
       \begin{bmatrix} 1 & A-0.5 \\ A-0.5 & 1
       \end{bmatrix}).
    \end{aligned}
    \end{equation}
Finally, based on Assumption \ref{assum: equ bias}, the short-term outcome and the long-term outcome are generated by:
   \begin{equation} \small
    \begin{aligned}
       & S = 1 + A + X + 2 A\times X + 0.5  X^2 + A \times  X^2 + U + \epsilon_s,
       \\&
       Y = 2 + 3 A + X + 4 A\times X +  X^2 + 2 A \times X^2 + 2\times U - S + \epsilon_y,
    \end{aligned}
    \end{equation}
where $\epsilon_s$ and $\epsilon_y$ are all Gaussian noises. 

As a result, all nuisance functions and HLCE have the following parametric forms:
\begin{equation} \small \label{app: nuisance forms}
    \begin{aligned}
      &  \mu_S^E(A,X) = 1+X+0.5X^2+A+2A\times X + A\times X^2, \\
      &  \mu_S^O(A,X) = 1+0.5X+0.5X^2+A+3A\times X +A\times X^2 , \\
      &  \mu_Y^O(A,X) = 1-0.5X+0.5X^2+ 2 A + 2 A \times X + A\times X^2, \\
      &  \pi^E(X) = \frac{1}{1+\exp(x)}, \\
      &  \pi^O(X) = \frac{1}{1+\exp(x)}, \\
      &  \pi^G(X) = p(G=E) ,\\
      &  \tau (X) = 2+2X+X^2.
    \end{aligned}
\end{equation}

\textbf{Dataset 2:} Following \cite{frauenestimating}, we simulate some of nuisance functions from Gaussian processes using the prior induced by Matern kernel \cite{Ras2005GPfML} as follows:
\begin{equation} \small \label{app: matern kernel}
    \begin{aligned}
      K_{l,\nu}(x_i,x_j)= 
      & \frac{1}{\Gamma(\nu)2^{\nu-1}} \left( \frac{\sqrt{2\nu}}{l} \| x_i-x_j \|_2 \right)^\nu
      \\ & \times K_\nu \left(  \frac{\sqrt{2\nu}}{l} \| x_i-x_j \|_2\right) ,
    \end{aligned}
\end{equation}
where $\Gamma$ is the Gamma function, $K_\nu$ is the modified Bessel function of second kind, $l$ is the length scale of the kernel, and $\nu$ controls the smoothness of the sampled functions. In our simulation, we set $l=1$ and $\nu=2$ for the nuisance functions. We denote $f_i \sim \mathcal{GP}(0,K_{l,\nu})$ for $i=0,1$.
The generation of treatments $A$, observed $X$, and unobserved $U$ are the same as dataset 1. Then, we generate the short-term and long-term outcomes as
   \begin{equation} \small
    \begin{aligned}
       & S = f_0(x) + A + 2 A\times X  + A \times  X^2 + U + \epsilon_s,
       \\&
       Y =  f_1(x) + 3 A + X + 4 A\times X +  2 A \times X^2 + 2\times U - S + \epsilon_y,
    \end{aligned}
    \end{equation}
where $\epsilon_s$ and $\epsilon_y$ are all Gaussian noises. 
This results in that $\mu_S^E$, $\mu_S^O$, and $\mu_S^Y$ are $2$-smooth functions, and the HLCE are much smoother since $\tau(X)=2+2X+X^2$. 

\textbf{IHDP Datasets:} Our data generation process is greatly inspired by the original one in \cite{hill2011bayesian} in the traditional setting. We reuse the original covariate of $25$ dimensions in the IHDP dataset as $\{ X,U \}$, and we randomly select its $8$ dimensions as $U$ and the rest as $X$. Then, we sample binary treatment $A$ and group indicator $G$ from $ G \sim \text{Bernoulli}(p_g) $,  $A|G=E \sim  \text{Bernoulli}(p_e)$, and $A|G=O \sim \text{Bernoulli}(p_o) $ where 
   \begin{equation} \small
    \begin{aligned}
       & p_g = \frac{1}{1+\exp(- X W_g + offset_g)}, 
       \\&
       p_e = \frac{1}{1+\exp(-X W_e  + offset_e)},
      \\&
       p_o = \frac{1}{1+\exp(- X W_{o,x} - 3 \times U W_{o,u} + offset_o)}, 
       \\&
        W_g, W_e, W_{o,x} \sim \mathcal{N}(\mathbf{0},I_x), 
        \quad  W_{o,u} \sim \mathcal{N}(\mathbf{0},I_u),
    \end{aligned}
    \end{equation}
where
$offset_g$ controls the sample proportion between observational data and experimental data as about $2:1$, and $offset_e$ and $offset_o$ are set to ensure the $p_e$ and $p_o$ are between $0.05$ and $0.95$.
Here $I_x$ and $I_u$ are $17 \times 17$ and $8 \times 8$ identity matrices respectively. Then we generate the outcomes as follows:
\begin{equation} \small
\begin{aligned}
   & S = 
   \begin{cases}
     X W_{s1,x} + 4 + U W_{u} +\epsilon_s & \text{if } A = 1, \\
     \exp ( (X +0.5) W_{s0,x} ) + U W_{u} +\epsilon_s & \text{if } A = 0.
\end{cases}
,
   \\&
   Y =    \begin{cases}
     X W_{y1,x} + 8 + 2 U W_{u}  - S +\epsilon_y & \text{if } A = 1, \\
     \exp ( (X +0.5) W_{y0,x} ) + 2 U W_{u} - S +\epsilon_y & \text{if } A = 0.
\end{cases},
\end{aligned}
\end{equation}
where $\epsilon_y$ and $\epsilon_s$ are Gaussian noises.
Here, $W_{s1,x}$, $W_{s0,x}$, $W_{y1,x} $, and $W_{y0,x} $ are of $17\times 1$ dimensions and $W_{u}$ is of $8\times 1$ dimensions, and their elements are sampled independently from $\{ 0, 0.1, 0.2, 0.3, 0.4 \}$ with probabilities $\{ 0.6, 0.1,0.1,0.1 \}$.

\textbf{News Dataset:} We reuse the original covariates in News dataset \cite{johansson2016learning} and divide them into observed $X$ of $332$ dimensions and unobserved $U$ of $166$ dimensions. Then we generate treatment $T$ and group indicator $G$ from $ G \sim \text{Bernoulli}(p_g) $,  $A|G=E \sim  \text{Bernoulli}(p_e)$, and $A|G=O \sim \text{Bernoulli}(p_o) $ where 
   \begin{equation} \small
    \begin{aligned}
       & p_g = \frac{1}{1+\exp( X V_1 + offset_g)}, 
       \\&
       p_e = \frac{1}{1+\exp(X V_2  + offset_e)},
      \\&
       p_o = \frac{1}{1+\exp( X V_3 + U V_4 + offset_o)}, 
    \end{aligned}
    \end{equation}
in which for $i=1,2,3,4$, $V_i=\tilde V_i/\overline{\tilde V_i} $ where $ \tilde V_i \sim \mathcal N (\mathbf{1},\mathbf{I})$ and $\overline{\tilde V_i}$ is the mean of $\tilde V_i$.
Here, $offset_g$ is set to ensure the proportion of the experimental data and observational data is about $1:4$, and $offset_e$ and $offset_o$ are set to ensure the $p_e$ and $p_o$ are between $0.05$ and $0.95$. Then we generate the short-term and long-term outcomes as follows:
\begin{equation} \small
\begin{aligned}
   & S = 
   \begin{cases}
     X V_5 + X^2 V_5 + U V_6 +\epsilon_s & \text{if } A = 1, \\
     2 X V_7 + 3 X^2 V_7 + U V_6 +\epsilon_s & \text{if } A = 0.
\end{cases}
,
   \\&
   Y =    \begin{cases}
     X V_8 + X^2 V_8 + 4 + 2 U V_6  - S +\epsilon_y & \text{if } A = 1, \\
     2 X V_9 + 3 X^2 V_9 + 2 U V_6  - S +\epsilon_y & \text{if } A = 0.
\end{cases},
\end{aligned}
\end{equation}
where $\epsilon_s$ and $\epsilon_y$ are Gaussian noises, and for $i=5,6,7,8,9$, $V_i \sim \mathcal N (\mathbf{1},\mathbf{I})$.
    
\subsection{Parametric MR Estimator}
\label{app: parametric mr}
In Section \ref{sec: exp}, to answer RQ1 (can $\hat \tau_{mr}(x)$ achieve multiple robustness?), we run our method $\hat \tau_{mr}(x)$ in dataset 1 as described in Section \ref{app: dgp}. As listed in Eq. \eqref{app: nuisance forms}, the nuisance functions have their specific parametric forms, thus we can use correctly or incorrectly specified parametric regression methods to verify the multiple robustness property.
To be clear, we restate the ground true forms of nuisance functions as follows:
\begin{equation} \small \label{app: forms}
    \begin{aligned}
      &  \mu_S^E(A,X) = 1+X+0.5X^2+A+2A\times X + A\times X^2, \\
      &  \mu_S^O(A,X) = 1+0.5X+0.5X^2+A+3A\times X +A\times X^2 , \\
      &  \mu_Y^O(A,X) = 1-0.5X+0.5X^2+ 2 A + 2 A \times X + A\times X^2, \\
      &  \pi^E(X) = \frac{1}{1+\exp(x)}, \\
      &  \pi^O(X) = \frac{1}{1+\exp(x)}, \\
      &  \pi^G(X) = p(G=E) .
    \end{aligned}
\end{equation}

In our experiments, for \textbf{correctly specified models}, we use the polynomial regression method with $2$ degree to fit nuisance functions $\mu_S^E(1,X)$, $\mu_S^E(0,X)$, $\mu_S^O(0,X) $, $\mu_S^O(1,X) $, $\mu_Y^O(0,X)$ , and $\mu_Y^O(1,X) $. To fit $ \pi^E(X) $ and $ \pi^O(X) $ we use logistic regression, and for $\pi^G(X)$ we use the sample frequency of $G=E$.

For \textbf{misspecified models}, we use the linear regression method to fit nuisance functions $\mu_S^E(1,X)$, $\mu_S^E(0,X)$, $\mu_S^O(0,X) $, $\mu_S^O(1,X) $, $\mu_Y^O(0,X)$ , and $\mu_Y^O(1,X) $. We also use the sample frequency of $G=O$ for $\pi^G(X)$ and a functional form $f(X) = \frac{1}{1+\exp(\alpha X^2)}$ (where $\alpha$ is a fitting parameter) to model $\pi^E(X)$ and $\pi^O(X)$.

\section{Proof of Theorem \ref{theo: identifi}}\label{app: proof of iden}
\begin{proof}
    \begin{equation} \label{eq: tau decom} \small
        \begin{aligned}
            & \tau(X)  \\
            = & \mathbb E[Y(1)|X,G=O] - \mathbb E[Y(0)|X,G=O] \\
            = & \mathbb E[Y(1)|X,G=O,A=1]p(A=1|X,G=O) 
                \\
              & + \mathbb E[Y(1)|X,G=O,A=0]p(A=0|X,G=O) \\
              & - \mathbb E[Y(0)|X,G=O,A=1]p(A=1|X,G=O) 
                \\
              & + \mathbb E[Y(0)|X,G=O,A=0]p(A=0|X,G=O) \\
            = & \mathbb E[Y(1)|X,G=O,A=1]p(A=1|X,G=O) 
                \\
              & + \mathbb E[Y(1)|X,G=O,A=0]p(A=0|X,G=O) \\
              & - \mathbb E[Y(0)|X,G=O,A=1]p(A=1|X,G=O) 
                \\
              & + \mathbb E[Y(0)|X,G=O,A=0]p(A=0|X,G=O) \\
              & + \mathbb E[Y(1)|X,G=O,A=1]p(A=0|X,G=O)
                \\
              & - \mathbb E[Y(1)|X,G=O,A=1]p(A=0|X,G=O) \\
              & + \mathbb E[Y(0)|X,G=O,A=0]p(A=1|X,G=O)
                \\
              & - \mathbb E[Y(0)|X,G=O,A=0]p(A=1|X,G=O) \\
            = & \mathbb E[Y(1)|X,G=O,A=1] - \mathbb E[Y(0)|X,G=O,A=0] \\
              & + \{ \mathbb E[Y(1)|X,G=O,A=0] - \mathbb E[Y(1)|x,G=O,A=1]\}
              \\
              & \times p(A=0|X,G=O) \\
              & + \{ \mathbb E[Y(0)|X,G=O,A=0] - \mathbb E[Y(0)|x,G=O,A=1]\}\\
              & \times p(A=1|X,G=O) \\
            = & \mathbb E[Y|X,G=O,A=1] - \mathbb E[Y|X,G=O,A=0] \\
              & + \{ \mathbb E\left[ S(1) |X,G=O,A=0\right] - \mathbb E\left[ S(1)  |X,G=O,A=1\right] \}\\
              & \times p(A=0|X,G=O) \\
              & + \{ \mathbb E\left[ S(0) |X,G=O,A=0\right] - \mathbb E\left[ S(0) |X,G=O,A=1\right]  \}\\
              & \times p(A=1|X,G=O) , 
        \end{aligned}
    \end{equation}
    where the first equality is based on Assumption \ref{assum: external validity of exp} and the last equality is based on Assumption \ref{assum: equ bias}.
    Similarly, for short-term conditional causal effects, we have:
    \begin{equation} \label{eq: stce decom}
        \begin{aligned}
            & \mathbb E[S(1)|X,G=O] - \mathbb E[S(0)|X,G=O] \\
            = & \mathbb E[S|X,G=O,A=1] - \mathbb E[S|X,G=O,A=0] \\
              & + \{ \mathbb E[S(1)|X,G=O,A=0] - \mathbb E[S(1)|x,G=O,A=1]\}\\
              & \times p(A=0|X,G=O) \\
              & + \{ \mathbb E[S(0)|X,G=O,A=0] - \mathbb E[S(0)|x,G=O,A=1]\}\\
              &\times p(A=1|X,G=O) \\
        \end{aligned}
    \end{equation}
    Then, combining Eq. \eqref{eq: tau decom} and \eqref{eq: stce decom}, we have
    \begin{equation} 
        \begin{aligned}
            \tau(X) 
            = & \mathbb E[Y|X,G=O,A=1] - \mathbb E[Y|X,G=O,A=0] \\
              & + \mathbb E[S(1)|X,G=O] - \mathbb E[S(0)|X,G=O] \\
            & - \mathbb E[S|X,G=O,A=1] + \mathbb E[S|X,G=O,A=0] \\
            = & \mathbb E[Y|X,G=O,A=1] - \mathbb E[Y|X,G=O,A=0] \\
              & + \mathbb E[S(1)|X,G=E] - \mathbb E[S(0)|X,G=E] \\
            & - \mathbb E[S|X,G=O,A=1] + \mathbb E[S|X,G=O,A=0] \\
            = & \mathbb E[Y|X,G=O,A=1] - \mathbb E[Y|X,G=O,A=0] \\
              & + \mathbb E[S|X,G=E,A=1] - \mathbb E[S|X,G=E,A=0] \\
            & - \mathbb E[S|X,G=O,A=1] + \mathbb E[S|X,G=O,A=0],
        \end{aligned}
    \end{equation}
    where the second equality is based on Assumption \ref{assum: external validity of exp} and the last equality is based on Assumption \ref{assum: internal validity of exp}.
\end{proof}

\section{Proof of Lemma \ref{lem: baselines consis}} \label{app: proof of consi}


\begin{proof}
   We first prove the consistency of $\hat \tau_{reg}(x)$:
\begin{equation} \small \label{prf: cons.reg.ab}
\begin{aligned}
   & \mathbb E\left[\hat Y_{reg}|X=x\right] \\
= & \mathbb E\left[\mathbb I (G=O) \left[ (-1)^{1-A} \left( Y - \hat \mu_Y^O(1-A,X)  \right.\right.\right. \\
    & \left. \left. \left. - S + \hat \mu_S^O(1-A,X) \right)  +\hat \mu_S^E(1,X) - \hat \mu_S^E(0,X) \right]  |X=x \right]  \\
    & +\mathbb E\left[ \mathbb I (G=E) \left[ (-1)^{1-A} \left(S - \hat \mu_S^E(1-A,X) \right) \right. \right. \\
    &  \left.\left.  +\hat  \mu_Y^O(1,X) - \hat \mu_Y^O(0,X) + \hat \mu_S^O(0,X) - \hat \mu_S^O(1,X)) \right] |X=x \right] \\
= & a_{reg} + b_{reg} 
\end{aligned}
\end{equation}
where we have
\begin{equation} \small \label{prf: cons.reg.a}
\begin{aligned}
   & a_{reg} \\
= & \mathbb E\left[\mathbb I (G=O) \left[ (-1)^{1-A} \left( Y - \hat \mu_Y^O(1-A,X)   
   \right. \right.\right. \\
    & \left.\left. \left. - S + \hat \mu_S^O(1-A,X) \right)  +\hat \mu_S^E(1,X) - \hat \mu_S^E(0,X) \right] |X=x \right]  \\
= &   (1-\pi^G(x))\mathbb E\left[ (-1)^{1-A} \left( Y - \hat \mu_Y^O(1-A,X)   
   \right. \right. \\
    & \left. \left. - S + \hat \mu_S^O(1-A,X) \right)  +\hat \mu_S^E(1,X) - \hat \mu_S^E(0,X) |G=O, X=x \right] \\
= &  \pi^O(x) (1-\pi^G(x))\mathbb E\left[ (-1)^{1-A} \left( Y - \hat \mu_Y^O(1-A,X)   - S
   \right. \right. \\
    & \left. \left.  + \hat \mu_S^O(1-A,X) \right)  +\hat \mu_S^E(1,X) - \hat \mu_S^E(0,X) |A=1, G=O, X=x \right] \\
    & + (1-\pi^O(x)) (1-\pi^G(x))\mathbb E\left[ (-1)^{1-A} \left( Y - \hat \mu_Y^O(1-A,X)  - S
   \right. \right. \\
    & \left. \left.  + \hat \mu_S^O(1-A,X) \right)  +\hat \mu_S^E(1,X) - \hat \mu_S^E(0,X) |A=0,G=O, X=x \right] \\
= &  \pi^O(x) (1-\pi^G(x))\mathbb E\left[  Y - \hat \mu_Y^O(0,X)  - S 
   \right. \\
    & \left.  + \hat \mu_S^O(0,X)   +\hat \mu_S^E(1,X) - \hat \mu_S^E(0,X) |A=1, G=O, X=x \right] \\
    & + (1-\pi^O(x)) (1-\pi^G(x))\mathbb E\left[  \hat \mu_Y^O(1,X) -Y + S   
   \right. \\
    & \left. - \hat \mu_S^O(1,X)  +\hat \mu_S^E(1,X) - \hat \mu_S^E(0,X) |A=0,G=O, X=x \right] \\
= &  \pi^O(x) (1-\pi^G(x))\mathbb E\left[  \mu_Y^O(1,X) - \hat \mu_Y^O(0,X)  -\mu_S^O(1,X) 
   \right. \\
    & \left.  + \hat \mu_S^O(0,X)   +\hat \mu_S^E(1,X) - \hat \mu_S^E(0,X) |A=1, G=O, X=x \right] \\
    & + (1-\pi^O(x)) (1-\pi^G(x))\mathbb E\left[  \hat \mu_Y^O(1,X) - \mu_Y^O(0,X) + \mu_S^O(0,X)   
   \right. \\
    & \left. - \hat \mu_S^O(1,X)  +\hat \mu_S^E(1,X) - \hat \mu_S^E(0,X) |A=0,G=O, X=x \right] \\
\end{aligned}
\end{equation}
and similarly we have
\begin{equation} \small \label{prf: cons.reg.b}
\begin{aligned}
   & b_{reg} \\
= & \mathbb E\left[  \mathbb I (G=E) \left[ (-1)^{1-A} \left(S - \hat \mu_S^E(1-A,X) \right)  \right.\right.  \\
    &  \left.\left. +\hat  \mu_Y^O(1,X) - \hat \mu_Y^O(0,X) + \hat \mu_S^O(0,X) - \hat \mu_S^O(1,X)) \right] |X=x \right] \\
= & \pi^E(x)\pi^G(x) \mathbb E   \left[ \mu_S^E(1,X) - \hat \mu_S^E(0,X) +\hat  \mu_Y^O(1,X)  \right. \\
    &  \left. - \hat \mu_Y^O(0,X) + \hat \mu_S^O(0,X) - \hat \mu_S^O(1,X)) |A=1,X=x,G=E \right] \\
    & +(1-\pi^E(x))\pi^G(x) \mathbb E   \left[  \hat \mu_S^E(1,X)- \mu_S^E(0,X)  + \hat  \mu_Y^O(1,X) \right. \\
    &  \left. - \hat \mu_Y^O(0,X) + \hat \mu_S^O(0,X) - \hat \mu_S^O(1,X)) |A=0,X=x,G=E \right] .
\end{aligned}
\end{equation}
Hence, by substituting the consistency results, i.e., $\hat \mu_S^E(a,x)=\mu_S^E(a,x), \hat \mu_S^O(a,x)=\mu_S^O(a,x), \hat \mu_Y^O(a,x)=\mu_Y^O(a,x)$ into Eq. \eqref{prf: cons.reg.a}, Eq. \eqref{prf: cons.reg.b} and Eq. \eqref{prf: cons.reg.ab}, we have:
\begin{equation} \small 
\begin{aligned}
   & \mathbb E\left[\hat Y_{reg}|X=x\right] \\
= &  \mu_Y^O(1,x) - \mu_Y^O(0,x) + \mu_S^E(1,x) - \mu_S^E(0,x) + \mu_S^O(0,x) - \mu_S^O(1,x)
\end{aligned}
\end{equation}
which is our desired result.

Next, we prove the consistency of $\hat \tau_{pro}(x)$:
\begin{equation} \small \label{prf: cons.pro.ab}
\begin{aligned}
   & \mathbb E\left[ \hat Y_{pro}|X=x\right] \\
= &  \mathbb E \left[  \frac{(-1)^{1-A}}{1-A+(-1)^{1-A} \hat \pi ^E(X)} \frac{\mathbb I(G=E)}{p(G=O)}(\frac{1}{\hat \pi ^G(X)}-1)S |X=x\right] \\
    & + \mathbb E \left[  \frac{\mathbb I(G=O)}{p(G=O)}\frac{(-1)^{1-A}}{1-A+(-1)^{1-A} \hat \pi ^O(X)} (Y-S) |X=x\right] \\
= & a_{pro}+b_{pro}
\end{aligned}
\end{equation}
where we have
\begin{equation} \small \label{prf: cons.pro.a}
\begin{aligned}
   & a_{pro} \\
= &  \mathbb E \left[  \frac{(-1)^{1-A}}{1-A+(-1)^{1-A} \hat \pi ^E(X)} \frac{\mathbb I(G=E)}{p(G=O)}(\frac{1}{\hat \pi ^G(X)}-1)S |X=x\right] \\
= &  \pi^G(x) \mathbb E \left[  \frac{(-1)^{1-A}}{1-A+(-1)^{1-A} \hat \pi ^E(X)}\frac{1- \hat \pi^G(x)}{ (1-\pi^G(X))\hat \pi ^G(X)} \right. \\
 & \left. \times \mu_S^E(A,X) |G=E,X=x\right]  \\
= & \pi^E(x) \pi^G(x) \mathbb E \left[  \frac{1}{ \hat \pi ^E(X)}\frac{1- \hat \pi^G(X)}{ (1-\pi^G(X))\hat \pi ^G(X)}\right. \\
 & \left. \times \mu_S^E(1,X) |A=1,G=E,X=x\right] \\
    & +  (1- \pi^E(x)) \pi^G(x) \mathbb E \left[  \frac{-1}{1- \hat \pi ^E(X)}\frac{1- \hat \pi^G(X)}{ (1-\pi^G(X))\hat \pi ^G(X)} \right. \\
    & \left. \times \mu_S^E(0,X) |A=0,G=E,X=x\right] \\
= &  \frac{ \pi^E(x) (1- \hat \pi^G(x)) \pi^G(x) }{ \hat \pi ^E(x) (1-\pi^G(x)) \hat \pi ^G(x)} \mu_S^E(1,x) \\
& -  
\frac{(1- \pi^E(x)) (1- \hat \pi^G(x)) \pi^G(x) }{(1- \hat \pi ^E(x))  (1-\pi^G(x))\hat \pi ^G(x)}  \mu_S^E(0,x)
\end{aligned}
\end{equation}
and similarly, we have
\begin{equation} \small \label{prf: cons.pro.b}
\begin{aligned}
   & b_{pro} \\
= &  \mathbb E \left[  \frac{\mathbb I(G=O)}{p(G=O)}\frac{(-1)^{1-A}}{1-A+(-1)^{1-A} \hat \pi ^O(X)} (Y-S) |X=x\right] \\
= & \frac{\pi^O(x)}{ \hat \pi^O(x)} (\mu_Y^O(1,x)- \mu_S^O(1,x) ) - \frac{1-\pi^O(x)}{1- \hat \pi^O(x)} (\mu_Y^O(0,x)- \mu_S^O(0,x) ).
\end{aligned}
\end{equation}
Hence, by substituting the consistency results, i.e.,  $\hat \pi^E(a,x)=\pi^E(a,x) , \hat \pi^O(a,x)=\pi^O(a,x), \hat \pi^G(a,x)=\pi^G(a,x)$, into Eq. \eqref{prf: cons.pro.a}, Eq. \eqref{prf: cons.pro.b} and Eq. \eqref{prf: cons.pro.ab}, we have 
\begin{equation} \small 
\begin{aligned}
   & \mathbb E\left[\hat Y_{pro}|X=x\right] \\
= &  \mu_Y^O(1,x) - \mu_Y^O(0,x) + \mu_S^E(1,x) - \mu_S^E(0,x) + \mu_S^O(0,x) - \mu_S^O(1,x)
\end{aligned}
\end{equation}
which is our desired result.

\end{proof}

\section{Proof of Lemma \ref{lem: mr consis}} 
\label{app: proof of mr consi}

\begin{proof}
Rewrite $\mathbb E [\hat Y_{mr}|X=x]$
\begin{equation} \small
  \begin{aligned}
   & \mathbb E [\hat Y_{mr}|X=x] \\
        = & 
        \mathbb E \left[ 
        \frac{(-1)^{1-A}}{1-A+(-1)^{1-A} \hat \pi^E(X)} \frac{\mathbb I(G=E)}{p(G=O)}(S- \hat \mu^E_S(A,X) ) \right. \\
        & \times (\frac{1}{ \hat \pi^G(X)} -1)  + \frac{\mathbb I(G=O)}{p(G=O)} \left( \frac{(-1)^{1-A}}{1-A+(-1)^{1-A} \hat \pi^O(X)}\right. \\
        & \times (Y - \hat \mu^O_Y(A,X) - S +  \hat \mu^O_S(A,X)  )  \\
        &  + \hat \mu^O_Y(1,X) - \hat \mu^O_Y(0,X) 
        + \hat \mu^E_S(1,X) -\hat \mu^E_S(0,X) \\
        & \left. \left. + \hat \mu^O_S(0,X) - \hat \mu^O_S(1,X) \right) |X=x   \right] \\
        = &  
        \mathbb E \left[ 
        \frac{(-1)^{1-A}}{1-A+(-1)^{1-A} \hat \pi^E(x)} \frac{\mathbb I(G=E)}{p(G=O)}( S - \hat \mu^E_S(A,X) ) \right.
        \\  & \left. \times (\frac{1}{ \hat \pi^G(X)} -1)  |X=x \right]
        \\  &  + \mathbb E \left[ \frac{(-1)^{1-A}}{1-A+(-1)^{1-A} \hat \pi^O(x)} \frac{\mathbb I(G=O)}{p(G=O)}   ( Y - \hat \mu^O_Y(A,X) 
        \right. \\
        & \left. - S +  \hat \mu^O_S(A,X)  ) |X=x \right] 
         + \hat \mu^O_Y(1,x) - \hat \mu^O_Y(0,x) 
        + \hat \mu^E_S(1,x) 
        \\ & -\hat \mu^E_S(0,x)  + \hat \mu^O_S(0,x) - \hat \mu^O_S(1,x)\\
        = & a + b + c,
  \end{aligned}
\end{equation}
where we let 
\begin{equation} \small
  \begin{aligned}
        a = & 
         \mathbb E \left[ 
        \frac{(-1)^{1-A}}{1-A+(-1)^{1-A} \hat \pi^E(x)} \frac{\mathbb I(G=E)}{p(G=O)}( S - \hat \mu^E_S(A,X) ) \right. \\
        & \left. \times (\frac{1}{ \hat \pi^G(X)} -1) |X=x \right] ;    \\
  \end{aligned}
\end{equation}
\begin{equation} \small
  \begin{aligned}
        b = &  \mathbb E \left[  \frac{(-1)^{1-A}}{1-A+(-1)^{1-A} \hat \pi^O(x)} \frac{\mathbb I(G=O)}{p(G=O)}( Y - \hat \mu^O_Y(A,X) \right. \\
        & \left. - S +  \hat \mu^O_S(A,X)  ) |X=x \right] ; \\
  \end{aligned}
\end{equation}
\begin{equation} \small
  \begin{aligned}
        c = & \hat \mu^O_Y(1,x) - \hat \mu^O_Y(0,x) 
        +  \hat \mu^E_S(1,x) -\hat \mu^E_S(0,x) \\
        & + \hat \mu^O_S(0,x) - \hat \mu^O_S(1,x) .
  \end{aligned}
\end{equation}

For the first term $a$, we have 
\begin{equation} \small
    \begin{aligned}
        a = & 
        \mathbb E \left[ 
        \frac{(-1)^{1-A}}{1-A+(-1)^{1-A} \hat \pi^E(x)} \frac{\mathbb I(G=E)}{p(G=O)}( S - \hat \mu^E_S(A,X) )
        \right. \\ & \left. \times
        (\frac{1}{ \hat \pi^G(X)} -1) |X=x \right] \\
         = & 
        \mathbb E \left[ 
        \frac{A-(1-A)}{A\hat \pi^E(x)+ (1-A)(1-\hat \pi^E(x)) }\mathbb I(G=E)
        \right. \\ & \left. \times
        ( S - \hat \mu^E_S(A,X) )(\frac{1-\hat \pi^G(X)}{ \hat \pi^G(X) (1-\pi^G(X))}) |X=x \right] \\
        = & 
        \pi^G(x) \mathbb E \left[ 
        \frac{A-(1-A)}{A\hat \pi^E(x)+ (1-A)(1-\hat \pi^E(x)) }( S - \hat \mu^E_S(A,X) )
        \right. \\ & \left. \times
        (\frac{1-\hat \pi^G(X)}{ \hat \pi^G(X) (1-\pi^G(X))}) |X=x,G=E \right] \\
        = & 
        \pi^E(x)\pi^G(x) \mathbb E \left[ 
        \frac{1}{\hat \pi^E(x) }( S - \hat \mu^E_S(A,X) )
        \right. \\ & \left. \times
        (\frac{1-\hat \pi^G(X)}{ \hat \pi^G(X) (1-\pi^G(X))}) |X=x,G=E,A=1 \right] \\
        & + (1-\pi^E(x))\pi^G(x) \mathbb E \left[ 
        \frac{-1}{1-\hat \pi^E(x)}( S - \hat \mu^E_S(A,X) )
        \right. \\ & \left. \times
        (\frac{1-\hat \pi^G(X)}{ \hat \pi^G(X) (1-\pi^G(X))}) |X=x,G=E,A=0 \right] \\
        = & 
        \frac{\pi^E(x) \pi^G(x) (1-\hat \pi^G(x))}{\hat \pi^E(x) \hat \pi^G(x) (1-\pi^G(x))} ( \mu^E_S(1,x) - \hat \mu^E_S(1,x) )
         \\ &
         -  \frac{(1-\pi^E(x)) \pi^G(x)(1-\hat \pi^G(x))}{(1-\hat \pi^E(x)) \hat \pi^G(x) (1-\pi^G(x))} ( \mu^E_S(0,x) - \hat \mu^E_S(0,x) )    
    \end{aligned}
\end{equation}

For the second term $b$, we have 
\begin{equation} \small
    \begin{aligned}
        b = &  \mathbb E \left[  \frac{(-1)^{1-A}}{1-A+(-1)^{1-A} \hat \pi^O(X)} \frac{\mathbb I(G=O)}{p(G=O)}( Y - \hat \mu^O_Y(A,X) 
        \right. \\ & \left. 
        - S +  \hat \mu^O_S(A,X)  ) |X=x \right]  \\
        = & (1-\pi^G(x)) \mathbb E \left[  \frac{A-(1-A)}{A \hat \pi^O(X) + (1-A)(1-\hat \pi^O(X)) } \frac{1}{1-\pi^G(X)}
        \right. \\ & \left. \times
        ( Y - \hat \mu^O_Y(A,X) - S +  \hat \mu^O_S(A,X)  ) |X=x,G=O \right] \\
        = & \pi^O(x) (1-\pi^G(x)) \mathbb E \left[  \frac{1}{\hat \pi^O(X)} \frac{1}{1-\pi^G(X)}( Y - \hat \mu^O_Y(A,X) 
        \right. \\ & \left. 
        - S +  \hat \mu^O_S(A,X)  ) |X=x,G=O,A=1 \right] \\
        & + (1-\pi^O(x))(1-\pi^G(x)) \mathbb E \left[  \frac{-1}{1- \hat \pi^O(X)} \frac{1}{1-\pi^G(X)}
        \right. \\ & \left. \times
        ( Y - \hat \mu^O_Y(A,X) - S +  \hat \mu^O_S(A,X)  ) |X=x,G=O,A=0 \right] \\
        = &   \frac{\pi^O(x)}{\hat \pi^O(x)}( \mu^O_Y(1,x) - \hat \mu^O_Y(1,x) - \mu^O_S(1,x) +  \hat \mu^O_S(1,x)  ) 
         \\ & 
        - \frac{(1-\pi^O(x)) }{1- \hat \pi^O(x)}( \mu^O_Y(0,x) - \hat \mu^O_Y(0,x) - \mu^O_S(0,x) +  \hat \mu^O_S(0,x)  ) 
    \end{aligned}
\end{equation}
And then, combining terms $a$, $b$, and $c$, we obtain
\begin{equation} \small
  \begin{aligned}
   & \mathbb E [\hat Y_{mr}|X=x]\\
   = & a + b + c \\
   = &  \frac{\pi^E(x) \pi^G(x) (1-\hat \pi^G(x))}{ (\hat \pi^E(x)) \hat \pi^G(x) (1-\pi^G(x))} ( \mu^E_S(1,x) - \hat \mu^E_S(1,x) ) 
   \\ & 
   - \frac{(1-\pi^E(x)) \pi^G(x)(1-\hat \pi^G(x))}{(1-\hat \pi^E(x))  \hat \pi^G(x) (1-\pi^G(x))} ( \mu^E_S(0,x) - \hat \mu^E_S(0,x) ) \\
    & +  \frac{\pi^O(x)}{\hat \pi^O(x)}( \mu^O_Y(1,x) - \hat \mu^O_Y(1,x) - \mu^O_S(1,x) +  \hat \mu^O_S(1,x)  ) 
     \\ & 
     - \frac{(1-\pi^O(x)) }{1- \hat \pi^O(x)}( \mu^O_Y(0,x) - \hat \mu^O_Y(0,x) - \mu^O_S(0,x)  +  \hat \mu^O_S(0,x)  )  
       \\ &
       + \hat \mu^O_Y(1,x) - \hat \mu^O_Y(0,x) 
        +  \hat \mu^E_S(1,x) -\hat \mu^E_S(0,x)
     + \hat \mu^O_S(0,x) \\ & - \hat \mu^O_S(1,x)
    \end{aligned}
\end{equation}

When $ \mu^O_S(A,X), \mu^E_S(A,X),\mu^O_Y(A,X)$ is consistent, terms $a,b$ become zero, and then we have
\begin{equation} \small
    \begin{aligned}
     & \mathbb E [\hat Y_{mr}|X=x] \\
     =& \hat \mu^O_Y(1,x) - \hat \mu^O_Y(0,x) 
        +  \hat \mu^E_S(1,x) -\hat \mu^E_S(0,x) + \hat \mu^O_S(0,x) 
        \\ & - \hat \mu^O_S(1,x) \\
     =&  \mu^O_Y(1,x) - \mu^O_Y(0,x)  +   \mu^E_S(1,x) - \mu^E_S(0,x) +  \mu^O_S(0,x) 
     \\ & -  \mu^O_S(1,x) ,
    \end{aligned}
\end{equation}
which is consistent.

When $ \pi^E(x), \pi^O(x),  \pi^G(X)$ is consistent, then we have
\begin{equation} \small
    \begin{aligned}
   & \mathbb E [\hat Y_{mr}|X=x] \\
   = &  \mu^E_S(1,x) - \hat \mu^E_S(1,x)
    - (\mu^E_S(0,x) - \hat \mu^E_S(0,x) ) 
    +   \mu^O_Y(1,x) 
    \\ &
    - \hat \mu^O_Y(1,x)
    - \mu^O_S(1,x) +  \hat \mu^O_S(1,x)  
    -  (\mu^O_Y(0,x) - \hat \mu^O_Y(0,x)
    \\ &
    - \mu^O_S(0,x) +  \hat \mu^O_S(0,x))    
    + \hat \mu^O_Y(1,x) - \hat \mu^O_Y(0,x) 
        +  \hat \mu^E_S(1,x) 
        \\ & -\hat \mu^E_S(0,x) 
        + \hat \mu^O_S(0,x) - \hat \mu^O_S(1,x)\\
    = & \mu^E_S(1,x)  
    -  \mu^E_S(0,x)  +   \mu^O_Y(1,x)  - \mu^O_S(1,x) 
    -  \mu^O_Y(0,x) \\ & + \mu^O_S(0,x) ,
    \end{aligned}
\end{equation}
which is consistent.

When $ \mu^E_S(x,A),  \pi^O(x)$ is consistent, then we have
\begin{equation} \small
  \begin{aligned}
    & \mathbb E [\hat Y_{mr}|X=x]\\
    = &   \mu^O_Y(1,x) - \hat \mu^O_Y(1,x) - \mu^O_S(1,x) +  \hat \mu^O_S(1,x)  
     - ( \mu^O_Y(0,x) 
     \\ &  - \hat \mu^O_Y(0,x) - \mu^O_S(0,x) +  \hat \mu^O_S(0,x)  )   
     + \hat \mu^O_Y(1,x) - \hat \mu^O_Y(0,x) 
       \\ &  +  \hat \mu^E_S(1,x) -\hat \mu^E_S(0,x) + \hat \mu^O_S(0,x) - \hat \mu^O_S(1,x) \\
    = & \mu^O_Y(1,x)  - \mu^O_S(1,x)   
     -  \mu^O_Y(0,x) + \mu^O_S(0,x)  +  \hat \mu^E_S(1,x) \\ &  -\hat \mu^E_S(0,x)  \\
    = & \mu^O_Y(1,x)  - \mu^O_S(1,x)   
     -  \mu^O_Y(0,x) + \mu^O_S(0,x)  +   \mu^E_S(1,x) \\ &  - \mu^E_S(0,x) ,
    \end{aligned}
\end{equation}
which is consistent.

When $\pi^E(x), \mu^O_S(A,X), \mu^O_Y(A,X), \pi^G(x)$    is consistent, then we have
\begin{equation} \small
  \begin{aligned}
   & \mathbb E [\hat Y_{mr}|X=x]\\
    = & \mu^E_S(1,x) - \hat \mu^E_S(1,x) 
      - ( \mu^E_S(0,x) - \hat \mu^E_S(0,x) ) 
      + \hat \mu^O_Y(1,x) \\ &  - \hat \mu^O_Y(0,x) 
        +  \hat \mu^E_S(1,x) -\hat \mu^E_S(0,x) + \hat \mu^O_S(0,x) - \hat \mu^O_S(1,x) \\
    = & \mu^E_S(1,x) 
      -  \mu^E_S(0,x) 
      + \hat \mu^O_Y(1,x) - \hat \mu^O_Y(0,x) 
       + \hat \mu^O_S(0,x) \\ &  - \hat \mu^O_S(1,x) \\
    = & \mu^E_S(1,x) 
      -  \mu^E_S(0,x) 
      +  \mu^O_Y(1,x) -  \mu^O_Y(0,x) 
       +  \mu^O_S(0,x) \\ &  -  \mu^O_S(1,x) ,
    \end{aligned}
\end{equation}
which is consistent. 

Hence, we can conclude that as long as one of the sets above is consistent, our estimator is consistent.
\end{proof}

\section{Proof of Theorem \ref{thm: convergence rate}}
\label{app: proof of mr rate}

\begin{proof}
    We apply Theorem 1 in Kennedy et~al. \cite{kennedy2023towards}, yielding that
    \begin{equation} \label{app: eq tau} \small
    \begin{aligned}
       \hat \tau(x) - \tau(x) 
       = & (\hat \tau(x) - \tilde \tau(x)) + (\tilde \tau(x) - \tau(x))\\
       = &  (\hat \tau(x) - \tilde \tau(x)) +  O_p(R_n^*(x)) \\
       = &  \hat{\mathbb E}_n[\hat r(X)|X=x] + o_p(R_n^*(x))  +  O_p(R_n^*(x)) \\
    \end{aligned}
    \end{equation}
    where $\mathcal R_n^*(x)$ is the oracle risk of second stage regression. 
    We analyze the $\hat r(x)$:
    \begin{equation} \label{app: eq rx} \small
        \begin{aligned}
            & \hat r (x) \\
            = & \mathbb E[\hat Y_{mr} |X=x ] - \tau(x) \\
            = & \frac{\pi^E(x)}{\hat \pi^E(x) } \frac{\pi^G(x) (1-\hat \pi^G(x))}{ \hat \pi^G(x) (1-\pi^G(x))} ( \mu^E_S(1,x) - \hat \mu^E_S(1,x) ) 
            \\ &
            - \frac{1-\pi^E(x)}{1-\hat \pi^E(x)}\frac{\pi^G(x)(1-\hat \pi^G(x))}{ \hat \pi^G(x) (1-\pi^G(x))} ( \mu^E_S(0,x) - \hat \mu^E_S(0,x) ) \\
            & +  \frac{\pi^O(x)}{\hat \pi^O(x)}( \mu^O_Y(1,x) - \hat \mu^O_Y(1,x) - \mu^O_S(1,x) +  \hat \mu^O_S(1,x)  ) 
            \\ &
            - \frac{(1-\pi^O(x)) }{1- \hat \pi^O(x)}( \mu^O_Y(0,x) - \hat \mu^O_Y(0,x) - \mu^O_S(0,x) +  \hat \mu^O_S(0,x)  )  \\
            & + \hat \mu^O_Y(1,x) - \hat \mu^O_Y(0,x) 
                +  \hat \mu^E_S(1,x) -\hat \mu^E_S(0,x)
            \\ &
            + \hat \mu^O_S(0,x) - \hat \mu^O_S(1,x)
            - (\mu^O_Y(1,x) - \mu^O_Y(0,x) 
            \\ &
            +   \mu^E_S(1,x) - \mu^E_S(0,x) +  \mu^O_S(0,x) -  \mu^O_S(1,x) ) \\
            = & (\frac{\pi^E(x) \pi^G(x) (1-\hat \pi^G(x)) }{\hat \pi^E(x)  \hat \pi^G(x) (1-\pi^G(x))} - 1) ( \mu^E_S(1,x) - \hat \mu^E_S(1,x) ) \\
            & - (\frac{(1-\pi^E(x)) \pi^G(x)(1-\hat \pi^G(x)) }{ (1-\hat \pi^E(x))\hat \pi^G(x) (1-\pi^G(x))} -1) ( \mu^E_S(0,x) - \hat \mu^E_S(0,x) ) \\
            & +  \frac{\pi^O(x)-\hat \pi^O(x)}{\hat \pi^O(x)} ( \mu^O_Y(1,x) - \hat \mu^O_Y(1,x)  ) 
            \\ & 
            - \frac{ \hat \pi^O(x)-\pi^O(x)  }{1- \hat \pi^O(x)}( \mu^O_Y(0,x) - \hat \mu^O_Y(0,x) )  \\
            & -  \frac{\pi^O(x)- \hat \pi^O(x)}{\hat \pi^O(x)}( \mu^O_S(1,x) - \hat \mu^O_S(1,x) )
            \\ &
            + \frac{ \hat \pi^O(x)-\pi^O(x) }{1- \hat \pi^O(x)}( \mu^O_S(0,x) - \hat \mu^O_S(0,x) ). 
        \end{aligned}
    \end{equation}

    Since
    \begin{equation} \small
    \begin{aligned}
       & \pi^E(x) \pi^G(x) (1-\hat \pi^G(x)) - \hat \pi^E(x)  \hat \pi^G(x) (1-\pi^G(x))
      \\
      = & \pi^E(x) \pi^G(x) (1-\hat \pi^G(x)) - \hat \pi^E(x)  \hat \pi^G(x) (1-\hat \pi^G(x)) 
      \\&
      + \hat \pi^E(x)  \hat \pi^G(x) (1-\hat \pi^G(x)) - \hat \pi^E(x)  \hat \pi^G(x) (1-\pi^G(x)) \\
      = & \left(\pi^E(x) \pi^G(x) - \hat \pi^E(x)  \hat \pi^G(x) \right) (1-\hat \pi^G(x)) 
      \\&
      + \hat \pi^E(x) \hat \pi^G(x) (\pi^G(x)- \hat \pi^G(x) )  \\
      = & (\pi^E(x) \pi^G(x) -  \hat \pi^E(x) \pi^G(x) + \hat \pi^E(x) \pi^G(x) 
      \\&
      - \hat \pi^E(x)  \hat \pi^G(x)) (1-\hat \pi^G(x)) + \hat \pi^E(x)\hat \pi^G(x) (\pi^G(x)- \hat \pi^G(x) )  \\
      =  & (\pi^E(x)-  \hat \pi^E(x)) \pi^G(x)  (1-\hat \pi^G(x)) 
      \\& + \hat \pi^E(x) (\pi^G(x) - \hat \pi^G(x))  (1-\hat \pi^G(x)) 
      \\& + \hat \pi^E(x)\hat \pi^G(x) (\pi^G(x)- \hat \pi^G(x) )  \\
      = & (\pi^E(x)-  \hat \pi^E(x)) \pi^G(x)  (1-\hat \pi^G(x)) + \hat \pi^E(x) (\pi^G(x) - \hat \pi^G(x)) ,
    \end{aligned}
    \end{equation}
    we have
    \begin{equation} \small
    \begin{aligned}
    & \frac{\pi^E(x) \pi^G(x) (1-\hat \pi^G(x)) }{\hat \pi^E(x)  \hat \pi^G(x) (1-\pi^G(x))} - 1 \\
    = & \frac{(\pi^E(x)-  \hat \pi^E(x)) \pi^G(x)  (1-\hat \pi^G(x)) + \hat \pi^E(x) (\pi^G(x) - \hat \pi^G(x))}{\hat \pi^E(x)  \hat \pi^G(x) (1-\pi^G(x))}.
    \end{aligned}
    \end{equation}
    Then, under Assumption \ref{asmp:bounded}
    and applying inequality $(a+b)^2\leq 2 (a^2+b^2)$,
    we have
    \begin{equation}  \label{app: eq pi1} \small
        \begin{aligned}
           & \left( \frac{(\pi^E(x)-  \hat \pi^E(x)) \pi^G(x)  (1-\hat \pi^G(x)) + \hat \pi^E(x) (\pi^G(x) - \hat \pi^G(x))} {\hat \pi^E(x)  \hat \pi^G(x) (1-\pi^G(x))} \right)^2 \\
           \leq & \frac{2}{ (\hat \pi^E(x))^2 (\hat \pi^G(x))^2 (1-\pi^G(x))^2 } 
           \\ & \times
           ( (\pi^E(x)-  \hat \pi^E(x))^2 (\pi^G(x))^2  (1-\hat \pi^G(x))^2 
           \\ & 
           + (\hat \pi^E(x))^2 (\pi^G(x) - \hat \pi^G(x))^2
           ) \\
           \asymp & (\pi^E(x)-  \hat \pi^E(x))^2 + (\pi^G(x) - \hat \pi^G(x))^2 .
        \end{aligned}
    \end{equation}
    
    Similarly, we have
    \begin{equation} \small
        \begin{aligned}
           & (1-\pi^E(x)) \pi^G(x)(1-\hat \pi^G(x)) - (1-\hat \pi^E(x))\hat \pi^G(x) (1-\pi^G(x)) \\
           = & ( (1-\pi^E(x)) \pi^G(x) -  (1-\hat \pi^E(x))\hat \pi^G(x)) (1-\hat \pi^G(x)) 
           \\ & 
           (1-\hat \pi^E(x))\hat \pi^G(x) (\pi^G(x) - \hat \pi^G(x)) \\
          = &  (\hat  \pi^E(x) -\pi^E(x)) \pi^G(x)  (1-\hat \pi^G(x))     \\
          & + (1- \hat  \pi^E(x)) (\pi^G(x) -\hat \pi^G(x) ) 
        \end{aligned}
    \end{equation}
    and under Assumption \ref{asmp:bounded} and applying inequality $(a+b)^2 \leq 2(a^2 + b^2)$, we obtain
   \begin{equation} \label{app: eq pi2} \small
    \begin{aligned}
        & \left( \frac{(1-\pi^E(x)) \pi^G(x)(1-\hat \pi^G(x)) }{ (1-\hat \pi^E(x))\hat \pi^G(x) (1-\pi^G(x))} -1 \right) ^2 \\
        \leq &  \frac{2  }{ (1-\hat \pi^E(x))^2 (\hat \pi^G(x))^2 (1-\pi^G(x))^2 } 
        \\& \times (\hat  \pi^E(x) -\pi^E(x))^2 (\pi^G(x))^2  (1-\hat \pi^G(x))^2  
        \\& +  (1- \hat  \pi^E(x))^2 (\pi^G(x) -\hat \pi^G(x) )^2 \\   
        \asymp &  (\pi^E(x)-  \hat \pi^E(x))^2 + (\pi^G(x) - \hat \pi^G(x))^2 
    \end{aligned}
    \end{equation}

    Based on Eq. \eqref{app: eq rx} and inequalities \eqref{app: eq pi1} and \eqref{app: eq pi2}, and due to the independence $(\hat \pi^E(x) ,\hat \pi^G(x)  ) \Vbar \hat \mu_S^E(a,x)$ and $\hat \pi^O(x) \Vbar ( \hat  \mu_Y^O(a,x), \hat \mu_S^O(a,x))$ from sample splitting, we have 
    \begin{equation} \small
        \begin{aligned}
            & \hat\tau (x)-\tau(x) 
            \\ =& O_p( R^*(n)^2 + (r_{\pi^E}(n)+r_{\pi^G}(n)) r_{\mu_S^E}(n)
            \\ & +  r_{\pi^O}(n) (r_{\mu_Y^O}(n) + r_{\mu_S^O}(n)) ),
        \end{aligned}
    \end{equation}
    which finishes our proof.
    
    \end{proof}

\section{Poof of Theorem \ref{theo: mf convergence rate}}
\label{app: proof of mr rate under smooth}
    \begin{proof}
    Under Assumption \ref{asmp:smooth}, $\pi^E$ is $\alpha$-smooth, and $\| \pi^E - \pi \|_{w,2}=O_p(n^{-\frac{1}{2+d/\alpha}})$, and similarly for $\pi^O, \pi^G, \mu_S^E, \mu_S^O, \mu_Y^O$ and $\tau$. Based on Theorem \ref{thm: convergence rate}, we directly obtain
  \begin{equation} 
  \begin{aligned}
         & \hat \tau (x) -  \tau(x)  =  
        O_p(n^{-\frac{1}{2+d/\kappa}}
        + n^{-(\frac{1}{2+d/\alpha}+\frac{1}{2+d/\eta})}
        \\ &+  n^{-(\frac{1}{2+d/\gamma}+\frac{1}{2+d/\eta})}  
        + n^{-(\frac{1}{2+d/\zeta} +\frac{1}{2+d/\beta})} 
        + n^{-(\frac{1}{2+d/\delta}+\frac{1}{2+d/\beta})} ),
  \end{aligned}
    \end{equation}
    and the estimator is oracle efficient if
    \begin{equation} \small
        \begin{aligned}
             & \max \{n^{-(\frac{1}{2+d/\alpha}+\frac{1}{2+d/\eta})} 
              , n^{-(\frac{1}{2+d/\gamma}+\frac{1}{2+d/\eta})}  
             , n^{-(\frac{1}{2+d/\zeta}+\frac{1}{2+d/\beta})} ,
             \\ & n^{-(\frac{1}{2+d/\delta}+\frac{1}{2+d/\beta})} \}
             \leq n^{-\frac{1}{2+d/\kappa}} \\
              \iff &
             \min \{\frac{1}{2+d/\alpha}+\frac{1}{2+d/\eta} , 
             \frac{1}{2+d/\gamma}+\frac{1}{2+d/\eta} , \\ & 
             \frac{1}{2+d/\zeta}+\frac{1}{2+d/\beta}
             ,\frac{1}{2+d/\delta}+\frac{1}{2+d/\beta} \}
             \geq \frac{1}{2+d/\kappa} \\
              \iff &
             \frac{1}{\kappa} \geq 
             \max \{ 
             \frac{d^2-4\alpha\eta}{4d\alpha\eta + d^2\alpha+d^2\eta},
             \frac{d^2-4\gamma\eta}{4d\gamma\eta + d^2\gamma+d^2\eta},
             \\ &
             \frac{d^2-4\zeta\beta}{4d\zeta\beta + d^2\zeta+d^2\beta},
             \frac{d^2-4\delta\beta}{4d\delta\beta + d^2\delta+d^2\beta}
             \}
        \end{aligned}
    \end{equation}
    which finishes the proof.
\end{proof}

\section{Proof of Corollary \ref{coro: reg conv rate}}
\label{app: proof of baseline rate}


\begin{proof}
    We first prove the naive estimator:
    \begin{equation} \label{app: eq naive rate} \small
        \begin{aligned}
        &  \hat \tau_{naive}(x) - \tau(x) \\
        = & 
        \hat \mu_Y^O(1,x) - \hat \mu_Y^O(0,x) + \hat \mu_S^E(1,x) - \hat \mu_S^E(0,x) 
        \\& + \hat \mu_S^O(1,x) - \hat \mu_S^O(0,x)
       -  \mu_Y^O(1,x) +  \mu_Y^O(0,x)
        \\ & -  \mu_S^E(1,x) +  \mu_S^E(0,x) -  \mu_S^O(1,x) +  \mu_S^O(0,x),
        \end{aligned}
    \end{equation}
    and under Assumption \ref{asmp:smooth} that $\mu_S^E, \mu_S^O, \mu_Y^O$ are $\eta$-smooth,  $\delta$-smooth, and $\zeta$-smooth, respectively, we can directly obtain
    $\hat\tau (x)_{naive} -\tau(x) = O_p( n^{-\frac{1}{2+d/\eta}} + n^{-\frac{1}{2+d/\delta}} + n^{-\frac{1}{2+d/\zeta}}  )$.

    Next, we prove the rate of $ \hat\tau (x)_{reg}$ and$ \hat\tau (x)_{pro}$. Similar to $ \hat\tau (x)_{mr}$, we apply Theorem 1 in Kenney et~al. \cite{kennedy2023towards}, thus we only need to analyze terms $\hat r_{reg} = \mathbb E[\hat Y_{reg} |X=x ] - \tau(x)$ and  $\hat r_{pro} = \mathbb E[\hat Y_{pro} |X=x ] - \tau(x)$.

    For the term $\hat r_{reg} $, by combining Eq. \eqref{prf: cons.reg.a} and Eq. \eqref{prf: cons.reg.b} and substituting $\tau(x)= \mu_Y^O(1,x) -  \mu_Y^O(0,x) +  \mu_S^E(1,x) -  \mu_S^E(0,x) +  \mu_S^O(0,x) -  \mu_S^O(1,x)$ into $\hat r_{reg}$, we have
    \begin{equation} \small
        \begin{aligned}
        & \hat r_{reg} \\
        = &  \pi^O(x) (1-\pi^G(x)) \left(  \mu_Y^O(0,x) - \hat \mu_Y^O(0,x)   + \hat \mu_S^O(0,x) -  \mu_S^O(0,x)  \right. \\
            & \left. +\hat \mu_S^E(1,x) -\mu_S^E(1,x)  + \mu_S^E(0,x) - \hat \mu_S^E(0,x) \right) \\
        & + (1-\pi^O(x)) (1-\pi^G(x)) \left(  \hat \mu_Y^O(1,x) -\mu_Y^O(1,x) + \mu_S^O(1,x)   
           \right. \\
            &  \left. - \hat \mu_S^O(1,x)  +\hat \mu_S^E(1,x) - \mu_S^E(1,x) 
            + \mu_S^E(0,x) - \hat \mu_S^E(0,x)  \right) \\
        & + \pi^E(x)\pi^G(x)  \left( \mu_S^E(0,x) - \hat \mu_S^E(0,x) +\hat  \mu_Y^O(1,x) - \mu_Y^O(1,x)  \right. \\
            & + \mu_Y^O(0,x) - \hat \mu_Y^O(0,x) + \hat \mu_S^O(0,x) - \mu_S^O(0,x) + \mu_S^O(1,x)) 
            \\& \left. - \hat \mu_S^O(1,x) \right) 
        + (1-\pi^E(x))\pi^G(x)  \left(  \hat \mu_S^E(1,x) - \mu_S^E(1,x)  \right. \\
            &  + \hat  \mu_Y^O(1,x) - \mu_Y^O(1,x)   +  \mu_Y^O(0,x) - \hat \mu_Y^O(0,x) + \hat \mu_S^O(0,x) \\
            & \left. -\mu_S^O(0,x)   + \mu_S^O(1,x)) - \hat \mu_S^O(1,x) \right)   - 
        \left( \mu_Y^O(1,x) -  \mu_Y^O(0,x) \right. \\
            & \left. +  \mu_S^E(1,x) -  \mu_S^E(0,x) +  \mu_S^O(0,x) -  \mu_S^O(1,x) \right) \\
        =  &  (1-\pi^O(x)) (1-\pi^G(x)) \left(  \hat \mu_Y^O(1,x) -\mu_Y^O(1,x) \right) 
            \\& + \pi^G(x) \left(  \hat \mu_Y^O(1,x) -\mu_Y^O(1,x) \right) 
            \\& +  \pi^O(x) (1-\pi^G(x))  \left(  \mu_Y^O(0,x) - \hat \mu_Y^O(0,x) \right)
            \\& +  \pi^G(x)  \left(  \mu_Y^O(0,x) - \hat \mu_Y^O(0,x) \right)
            \\&  + (1-\pi^G(x)) \left( \hat \mu_S^E(1,x) -\mu_S^E(1,x)   \right)
            \\&  + (1-\pi^E(x))\pi^G(x) \left( \hat \mu_S^E(1,x) -\mu_S^E(1,x)   \right)
            \\&  + (1-\pi^G(x)) \left( \mu_S^E(0,x) - \hat \mu_S^E(0,x) \right)
            \\&  + \pi^E(x)\pi^G(x) \left( \mu_S^E(0,x) - \hat \mu_S^E(0,x) \right)
            \\&  +  \pi^O(x) (1-\pi^G(x)) \left( \hat \mu_S^O(0,x) -  \mu_S^O(0,x)\right) 
            \\& + \pi^G(x) \left( \hat \mu_S^O(0,x) -  \mu_S^O(0,x)\right) 
            \\& + (1-\pi^O(x)) (1-\pi^G(x)) \left(  \mu_S^O(1,x) - \hat \mu_S^O(1,x)\right)
            \\& + \pi^G(x) \left( \mu_S^O(1,x) - \hat \mu_S^O(1,x)\right).
        \end{aligned}
    \end{equation}
    By applying $ ( \Sigma_i^n a_i )^2 \leq n \Sigma_i^n a_i^2$ and under Assumption \ref{asmp:bounded}, we obtain
     \begin{equation} \small \label{app: eq reg rate}
        \begin{aligned}
        & \hat r_{reg}^2 \\
        \leq &\frac{1}{12}   \left[   (1-\pi^O(x))^2 (1-\pi^G(x))^2 \left(  \hat \mu_Y^O(1,x) -\mu_Y^O(1,x) \right)^2  \right.
            \\& + (\pi^G(x))^2 \left(  \hat \mu_Y^O(1,x) -\mu_Y^O(1,x) \right)^2 
            \\& +  (\pi^O(x))^2 (1-\pi^G(x))^2  \left(  \mu_Y^O(0,x) - \hat \mu_Y^O(0,x) \right)^2
            \\& +  (\pi^G(x))^2  \left(  \mu_Y^O(0,x) - \hat \mu_Y^O(0,x) \right)^2
            \\&  + (1-\pi^G(x))^2 \left( \hat \mu_S^E(1,x) -\mu_S^E(1,x)   \right)^2
            \\&  + (1-\pi^E(x))^2 (\pi^G(x))^2 \left( \hat \mu_S^E(1,x) -\mu_S^E(1,x)   \right)^2
            \\&  + (1-\pi^G(x))^2 \left( \mu_S^E(0,x) - \hat \mu_S^E(0,x) \right)^2
            \\&  + (\pi^E(x))^2 (\pi^G(x))^2 \left( \mu_S^E(0,x) - \hat \mu_S^E(0,x) \right)^2
            \\&  +  (\pi^O(x))^2 (1-\pi^G(x))^2 \left( \hat \mu_S^O(0,x) -  \mu_S^O(0,x)\right)^2 
            \\& + (\pi^G(x))^2 \left( \hat \mu_S^O(0,x) -  \mu_S^O(0,x)\right)^2 
            \\& + (1-\pi^O(x))^2 (1-\pi^G(x))^2 \left(  \mu_S^O(1,x) - \hat \mu_S^O(1,x)\right)^2
            \\& \left. + (\pi^G(x))^2 \left( \mu_S^O(1,x) - \hat \mu_S^O(1,x)\right)^2 \right] \\   
        \asymp &  (\hat \mu_Y^O(1,x) -\mu_Y^O(1,x))^2 
            + (\mu_Y^O(0,x) - \hat \mu_Y^O(0,x))^2 
            \\&+ (\hat \mu_S^E(1,x) -\mu_S^E(1,x))^2 
            + (\mu_S^E(0,x) - \hat \mu_S^E(0,x) )^2 
            \\&+ (\hat \mu_S^O(0,x) -  \mu_S^O(0,x) )^2 
            + (\mu_S^O(1,x) - \hat \mu_S^O(1,x) )^2 .
        \end{aligned}
    \end{equation}
    Similarly to $\hat \tau _{mr} (x)$, under Assumption \ref{asmp:smooth}, we can obtain
    \begin{equation} \small
      \begin{aligned}
        \hat\tau_{reg} (x) -\tau(x) = O_p(n^{-\frac{1}{2+d/\kappa}}+ n^{-\frac{1}{2+d/\eta}} + n^{-\frac{1}{2+d/\delta}} + n^{-\frac{1}{2+d/\zeta}}  ).
      \end{aligned}
    \end{equation}
    And $\hat\tau_{reg} (x)$ attain the oracle rate if
    \begin{equation} \small
        \begin{aligned}
             & \max \{n^{-\frac{1}{2+d/\eta}} + n^{-\frac{1}{2+d/\delta}} + n^{-\frac{1}{2+d/\zeta}} \}
             \leq n^{-\frac{1}{2+d/\kappa}} \\
              \iff &
             \min \{\frac{1}{2+d/\eta}, \frac{1}{2+d/\delta}, \frac{1}{2+d/\zeta}\}
             \geq \frac{1}{2+d/\kappa} \\
              \iff &
             \kappa \leq 
             \min \{ 
             \eta, \delta, \zeta
             \}
        \end{aligned}
    \end{equation}

    For the term $\hat r_{pro}$, by combining Eq. \eqref{prf: cons.pro.a} and Eq. \eqref{prf: cons.pro.b} and substituting $\tau(x)= \mu_Y^O(1,x) -  \mu_Y^O(0,x) +  \mu_S^E(1,x) -  \mu_S^E(0,x) +  \mu_S^O(0,x) -  \mu_S^O(1,x)$ into $\hat r_{pro}$, we have
    \begin{equation} \small
        \begin{aligned}
        &\hat r_{pro} \\
        = &  \frac{ \pi^E(x) (1- \hat \pi^G(x)) \pi^G(x) }{ \hat \pi ^E(x) (1-\pi^G(x)) \hat \pi ^G(x)} \mu_S^E(1,x) \\
        & -  \frac{(1- \pi^E(x)) (1- \hat \pi^G(x)) \pi^G(x) }{(1- \hat \pi ^E(x))  (1-\pi^G(x))\hat \pi ^G(x)}  \mu_S^E(0,x) \\    
        & + \frac{\pi^O(x)}{ \hat \pi^O(x)} (\mu_Y^O(1,x)- \mu_S^O(1,x) ) \\
        &  - \frac{1-\pi^O(x)}{1- \hat \pi^O(x)} (\mu_Y^O(0,x)- \mu_S^O(0,x) ) 
        - \left( \mu_Y^O(1,x) -  \mu_Y^O(0,x) \right. \\
        & \left. +  \mu_S^E(1,x) -  \mu_S^E(0,x) +  \mu_S^O(0,x) -  \mu_S^O(1,x) \right) \\
        = &  \left( \frac{ \pi^E(x) (1- \hat \pi^G(x)) \pi^G(x) }{ \hat \pi ^E(x) (1-\pi^G(x)) \hat \pi ^G(x)} -1 \right) \mu_S^E(1,x) \\
        & +  \left( 1- \frac{(1- \pi^E(x)) (1- \hat \pi^G(x)) \pi^G(x) }{(1- \hat \pi ^E(x))  (1-\pi^G(x))\hat \pi ^G(x)} \right) \mu_S^E(0,x) \\    
        & + \left( \frac{\pi^O(x)}{ \hat \pi^O(x)} -1 \right) (\mu_Y^O(1,x)- \mu_S^O(1,x) ) \\
        &  + \left( 1- \frac{1-\pi^O(x)}{1- \hat \pi^O(x)}  \right) (\mu_Y^O(0,x)- \mu_S^O(0,x) )  .
        \end{aligned}
    \end{equation}
    By applying $ ( \Sigma_i^n a_i )^2 \leq n \Sigma_i^n a_i^2$ and under Assumption \ref{asmp:bounded}, we obtain
    \begin{equation} \small \label{app: eq pro rate}
        \begin{aligned}
        &\hat r_{pro}^2 \\ 
        \leq & \frac{1}{4} \left[ \left( \frac{ \pi^E(x) (1- \hat \pi^G(x)) \pi^G(x) }{ \hat \pi ^E(x) (1-\pi^G(x)) \hat \pi ^G(x)} -1 \right)^2 (\mu_S^E(1,x)) ^2 \right. \\
        & +  \left( 1- \frac{(1- \pi^E(x)) (1- \hat \pi^G(x)) \pi^G(x) }{(1- \hat \pi ^E(x))  (1-\pi^G(x))\hat \pi ^G(x)} \right)^2 (\mu_S^E(0,x))^2 \\    
        & + \left( \frac{\pi^O(x)}{ \hat \pi^O(x)} -1 \right)^2 (\mu_Y^O(1,x)- \mu_S^O(1,x) )^2 \\
        & \left. + \left( 1- \frac{1-\pi^O(x)}{1- \hat \pi^O(x)} \right) ^2 (\mu_Y^O(0,x)- \mu_S^O(0,x) )^2  \right] \\
        \asymp & (\pi^E(x) -\hat \pi^E(x) )^2 + (\pi^G(x) - \hat \pi^G(x))^2  (\pi^O(x) - \hat \pi^O(x))^2
        \end{aligned}
    \end{equation}
     Similarly to $\hat \tau _{mr} (x)$ and $\hat \tau _{reg} (x)$, under Assumption \ref{asmp:smooth}, we can obtain
    \begin{equation} \small
      \begin{aligned}
        \hat\tau_{pro} (x) -\tau(x) = O_p(n^{-\frac{1}{2+d/\kappa}}+ n^{-\frac{1}{2+d/\alpha}} + n^{-\frac{1}{2+d/\beta}} + n^{-\frac{1}{2+d/\gamma}} ),
      \end{aligned}
    \end{equation}
    and $\hat\tau_{reg} (x)$ attain the oracle rate if
    \begin{equation} \small
        \begin{aligned}
             & \max \{n^{-\frac{1}{2+d/\alpha}} + n^{-\frac{1}{2+d/\beta}} + n^{-\frac{1}{2+d/\gamma}} \}
             \leq n^{-\frac{1}{2+d/\kappa}} \\
              \iff &
             \min \{\frac{1}{2+d/\alpha}, \frac{1}{2+d/\beta}, \frac{1}{2+d/\zeta}\}
             \geq \frac{1}{2+d/\gamma} \\
              \iff &
             \kappa \leq 
             \min \{ 
             \alpha, \beta, \gamma
             \}
        \end{aligned}
    \end{equation}
    which finishes our proof.

\end{proof}

\section{Proof of Theorem \ref{theo: rate under sparsity}}
\label{app: proof of sparsity rate}

\begin{proof}
    The proof follows immediately from the proofs of Theorem \ref{thm: convergence rate}, Theorem \ref{theo: mf convergence rate}, and Corollary \ref{coro: reg conv rate} by applying Assumption \ref{asmp:sparse}. 
    
    Specifically, from Eq. \eqref{app: eq naive rate} and Assumption  \ref{asmp:sparse}, we have 
    $\hat\tau_{naive} (x) -\tau(x) 
         = O_p( \frac{d_{\mu_S^E} \log (d)}{n} 
         + \frac{d_{\mu_S^O} \log (d)}{n} 
         + \frac{d_{\mu_Y^O} \log (d)}{n}  )$.
    From Eq. \eqref{app: eq pro rate}
     \begin{equation} \small 
        \begin{aligned}
        \hat r_{pro}^2  
        \asymp & (\pi^E(x) -\hat \pi^E(x) )^2 + (\pi^G(x) - \hat \pi^G(x))^2  (\pi^O(x) - \hat \pi^O(x))^2
        \end{aligned}
    \end{equation}
    and Assumption \ref{asmp:sparse}, we have $\hat\tau_{pro} (x) -\tau(x) 
        = O_p( \frac{d_{\tau} \log (d)}{n} 
        + \frac{d_{\pi^E} \log (d)}{n} 
        + \frac{d_{\pi^O} \log (d)}{n} 
        + \frac{d_{\pi^G} \log (d)}{n}   )
    $.
    From Eq. \eqref{app: eq reg rate}
     \begin{equation} \small 
        \begin{aligned}
         \hat r_{reg}^2
        \asymp &  (\hat \mu_Y^O(1,x) -\mu_Y^O(1,x))^2 
            + (\mu_Y^O(0,x) - \hat \mu_Y^O(0,x))^2 
            \\&+ (\hat \mu_S^E(1,x) -\mu_S^E(1,x))^2 
            + (\mu_S^E(0,x) - \hat \mu_S^E(0,x) )^2 
            \\&+ (\hat \mu_S^O(0,x) -  \mu_S^O(0,x) )^2 
            + (\mu_S^O(1,x) - \hat \mu_S^O(1,x) )^2 
        \end{aligned}
    \end{equation}
    and Assumption  \ref{asmp:sparse}, we have $\hat\tau_{reg} (x) -\tau(x) 
        = O_p( \frac{d_{\tau} \log (d)}{n}
        + \frac{d_{\mu_S^E} \log (d)}{n} 
        + \frac{d_{\mu_S^O} \log (d)}{n} 
        + \frac{d_{\mu_Y^O} \log (d)}{n}  )$.
    From Theorem \ref{thm: convergence rate} and Assumption \ref{asmp:sparse}, we have
    $\hat\tau _{mr} (x)-\tau(x) 
        =  O_p(  \frac{d_{\tau} \log (d)}{n}
         + \frac{(d_{\pi^E} + d_{\mu_S^E}) \log ^2(d)}{n^2} 
         + \frac{(d_{\pi^G} + d_{\mu_S^E}) \log ^2(d)}{n^2} 
         + \frac{(d_{\pi^O} + d_{\mu_Y^O}) \log ^2(d)}{n^2}
         + \frac{(d_{\pi^O} + d_{\mu_S^O}) \log ^2(d)}{n^2}  )$.
    Furthermore, the conditions under which these estimators achieve oracle efficiency can be directly obtained by comparing the oracle rate and the rest of the rate, similarly to Theorem \ref{theo: mf convergence rate} and Corollary \ref{coro: reg conv rate}.
    
\end{proof}

}

\bibliographystyle{IEEEtran}
\bibliography{IEEEabrv,ref}

\newpage

 




\begin{IEEEbiography}
[{\includegraphics[width=1in, height=1.25in, clip, keepaspectratio]{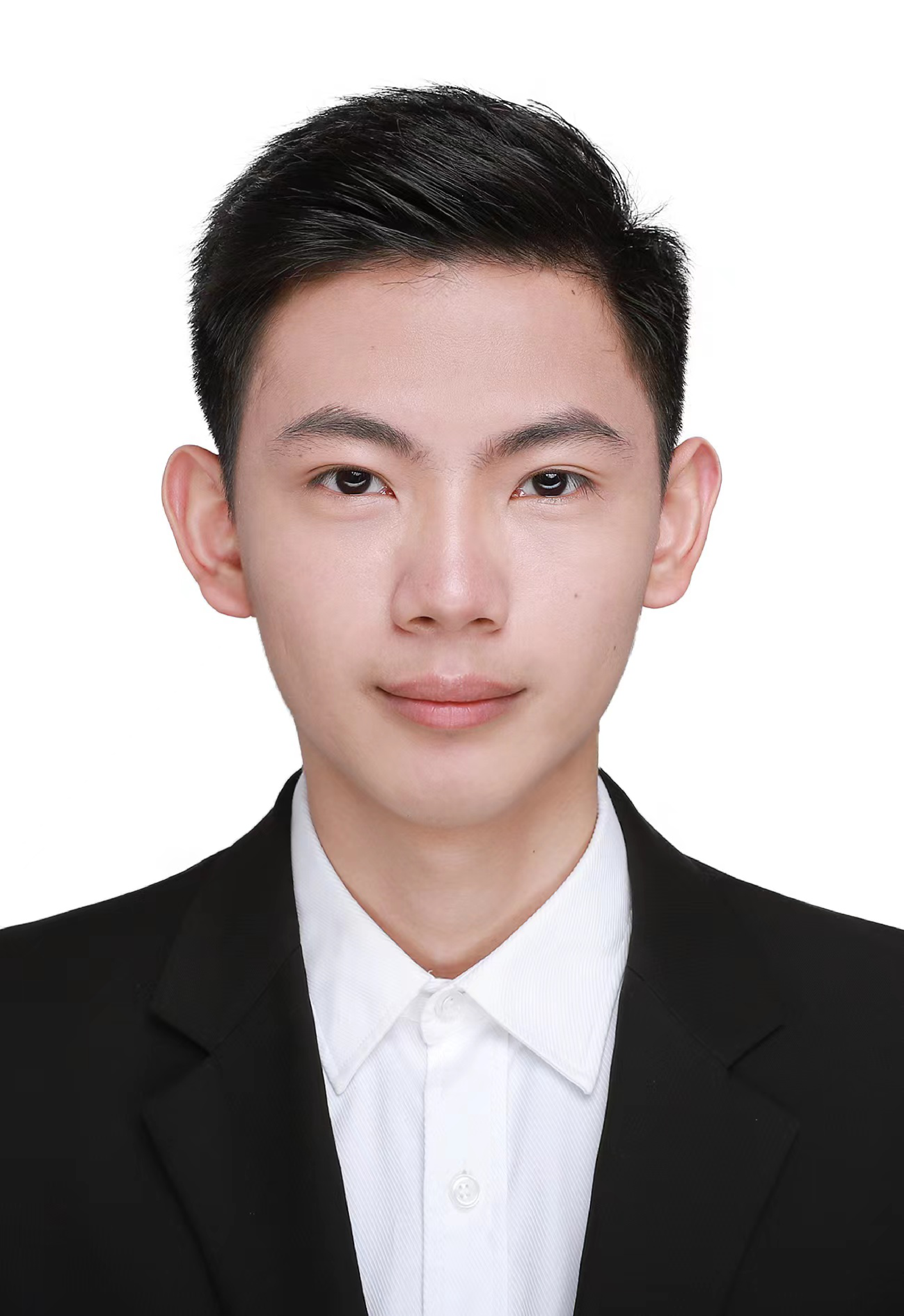}}]{Weilin Chen} received the B.S. degree in software engineering from Guangdong University of Technology, Guangzhou, China, in 2020, where he is currently pursuing the Ph.D. degree with the School of Computer. His current research interests include causal inference and machine learning.
\end{IEEEbiography}

\begin{IEEEbiography}[{\includegraphics[width=1in, height=1.25in, clip, keepaspectratio]{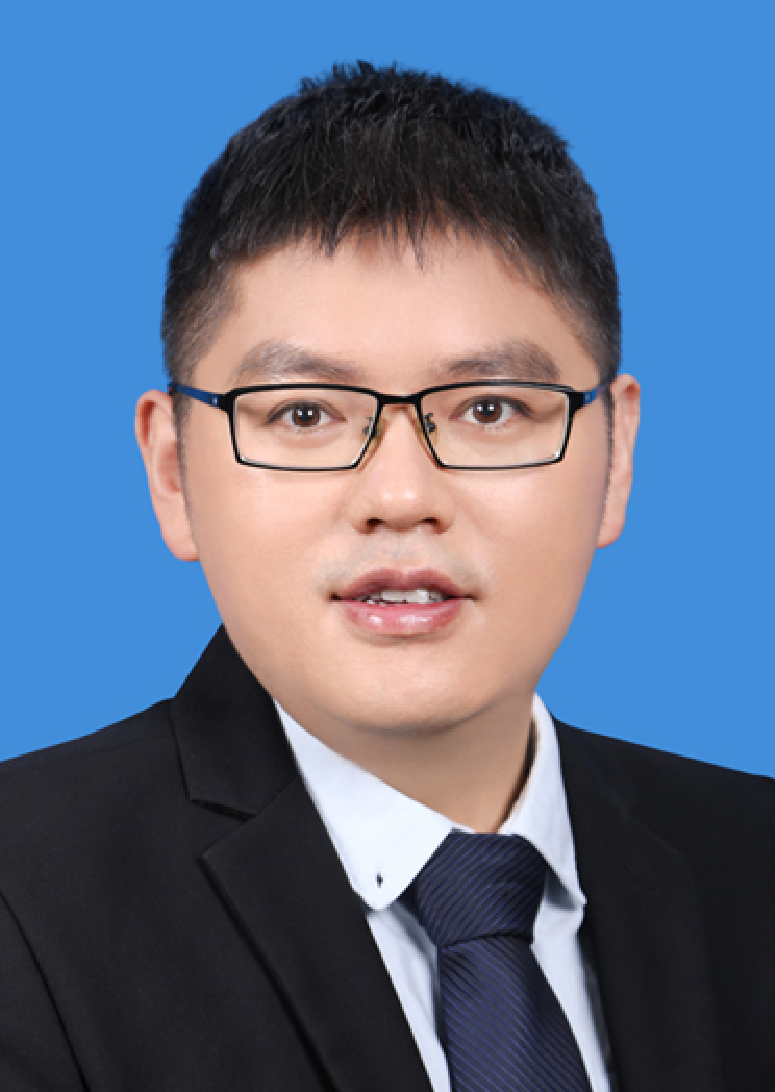}}]{Ruichu Cai} (M'17) is currently a professor in the school of computer science and the director of the data mining and information retrieval laboratory, Guangdong University of Technology. He received his B.S. degree in applied mathematics and Ph.D. degree in computer science from South China University of Technology in 2005 and 2010, respectively. 
 
 His research interests cover various topics, including causality, deep learning, and their applications. He was a recipient of the National Science Fund for Excellent Young Scholars, the Natural Science Award of Guangdong, and so on awards. He has served as the area chair of ICML 2022, NeurIPS 2022, and UAI 2022, senior PC for AAAI 2019-2022, IJCAI 2019-2022, and so on. He is now a senior member of CCF and IEEE.

\end{IEEEbiography}

\begin{IEEEbiography}
[{\includegraphics[width=1in, height=1.25in, clip, keepaspectratio]
{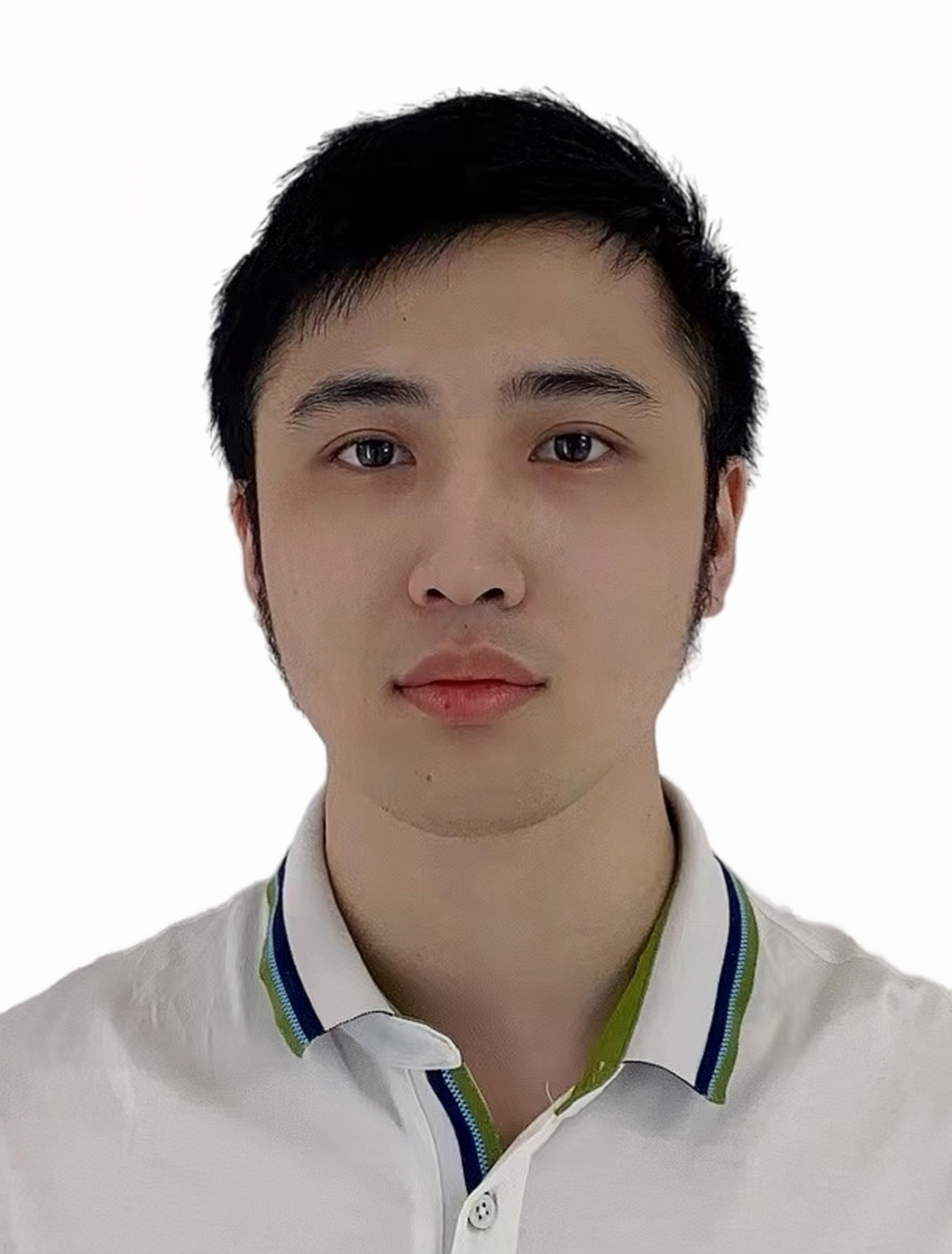}}]{Junjie Wan}  received the B.S. degree in computer science and technology from South China Agricultural University, Guangzhou, China, in 2021. Currently, he is pursuing the Master's degree at the School of Computer, Guangdong University of Technology. His current research interests include causal inference and machine learning.

\end{IEEEbiography}

\begin{IEEEbiography}
[{\includegraphics[width=1in, height=1.25in, clip, keepaspectratio]
{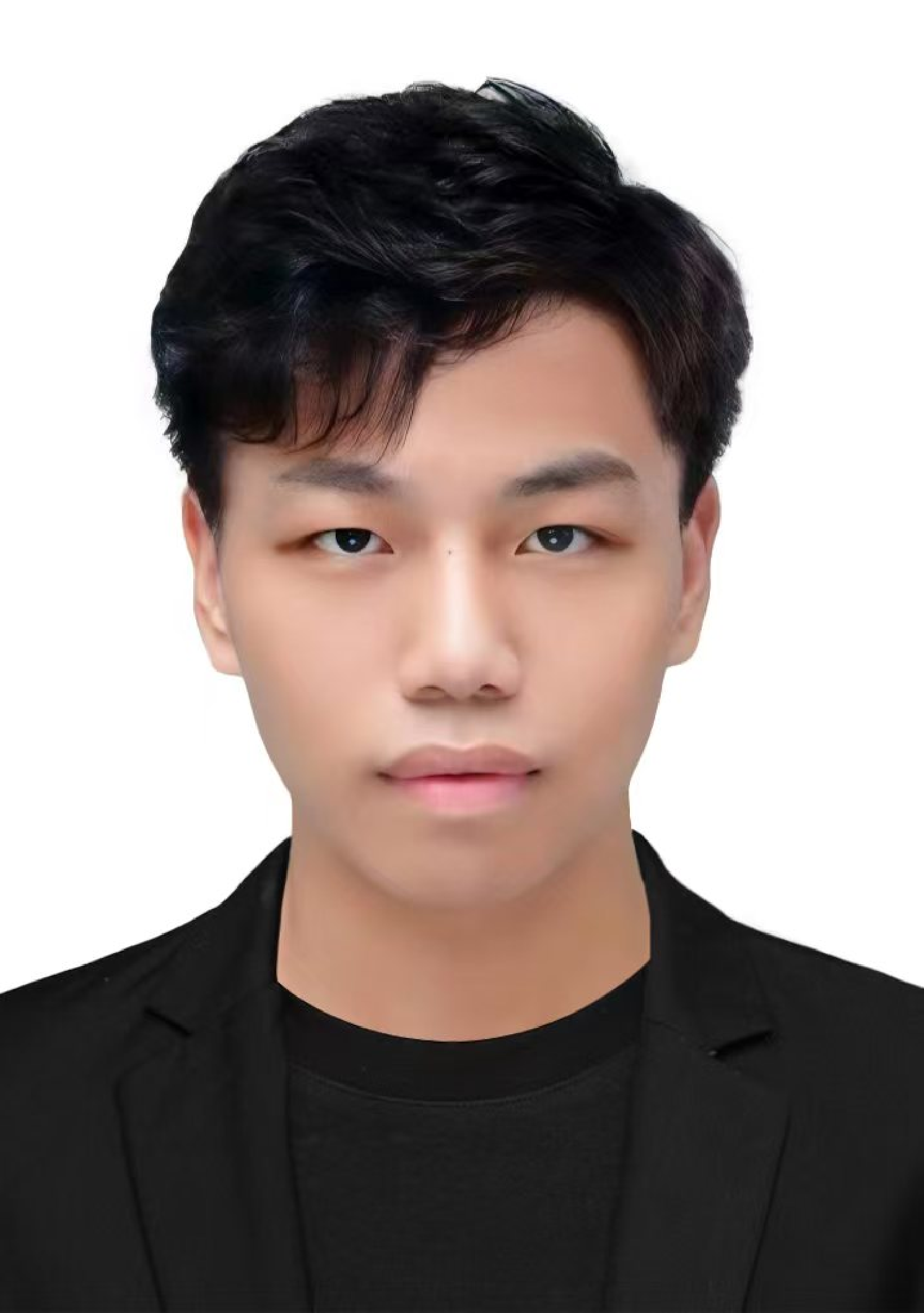}}]{Zeqin Yang}  received his B.S. degree in software
engineering from Guangdong University of Technology, Guangzhou, China, in 2022. 
He is now a Master's student at the School of Computer, Guangdong University of Technology. His current research interests lie in causal inference and its applications.

\end{IEEEbiography}

\begin{IEEEbiography}
[{\includegraphics[width=1in, height=1.25in, clip, keepaspectratio]
{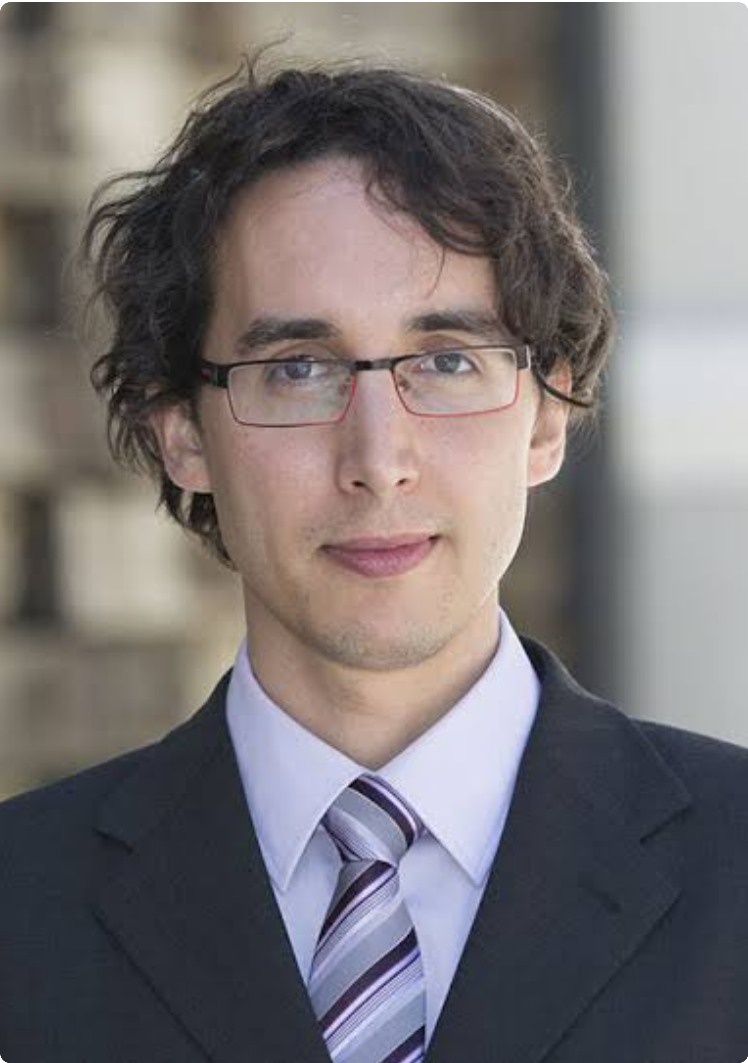}}]{Jos\'e Miguel Hern\'andez-Lobato} 
is Professor of Machine Learning at the Department of Engineering in the University of Cambridge, UK. Before joining Cambridge as faculty, he was a postdoctoral fellow at Harvard University, and before this, also a postdoctoral research associate at the University of Cambridge. Jose Miguel completed his Ph.D. and M.Phil. in Computer Science at Universidad Autónoma de Madrid (Spain), where he also obtained a B.Sc. in Computer Science from this institution, with a special prize to the best academic record on graduation. José Miguel's research interests are on probabilistic machine learning, with a focus on deep generative models, Bayesian optimization, approximate inference, causal inference, Bayesian neural networks and applications of these methods to real-world problems.

\end{IEEEbiography}


\vfill

\end{document}